\documentclass[twoside]{article}
\usepackage{ecj,palatino,epsfig,latexsym,natbib}
\usepackage{amsmath,amssymb,amsfonts}
\usepackage{algorithm}
\usepackage{algpseudocode}
\usepackage{graphicx}
\usepackage{xcolor}
\usepackage{booktabs}
\usepackage{url}

\begin{document}

\ecjHeader{x}{x}{xxx-xxx}{200X}
{MEMENTO for Code-as-Policy Evolution}
{A. Sygkounas, V. Aregbede, A. Loutfi, and A. Persson}

\title{\bf MEMENTO: Memory-Guided Memetic
Code-as-Policy Evolution}

\author{
\name{\bf Alkis Sygkounas} \hfill \addr{alkis.sygkounas@oru.se}\\
\addr{Center for Applied Autonomous Sensor Systems (AASS), {\"O}rebro University, Sweden}
\AND
\name{\bf Victor Aregbede} \hfill \addr{victor.aregbede@oru.se}\\
\addr{Center for Applied Autonomous Sensor Systems (AASS), {\"O}rebro University, Sweden}
\AND
\name{\bf Amy Loutfi} \hfill \addr{amy.loutfi@oru.se}\\
\addr{ Örebro University, 70281 Örebro, Sweden and
 Linköing University, 58183 Linköping, Sweden}
\AND
\name{\bf Andreas Persson} \hfill \addr{andreas.persson@oru.se}\\
\addr{Center for Applied Autonomous Sensor Systems (AASS), {\"O}rebro University, Sweden}
}
\maketitle

\begin{abstract}

Long-horizon embodied tasks require policies that execute many dependent actions before task success can be observed. Representing policies as executable control programs (code-as-policy) enables their decision logic to be inspected and revised after rollout evaluation. Revised programs can then be executed and compared by rollout performance, framing policy improvement as execution-guided program search. Evolutionary methods driven by large language models (LLMs) provide a natural mechanism for this search by generating variants and selecting high-performing candidates. However, existing approaches primarily select among independently generated variants and lack a sequential local improvement phase. We introduce \textsc{MEMENTO}, a memory-guided single-elite memetic framework for code-as-policy evolution. \textsc{MEMENTO} first evolves a rollout evaluator that maps policy rollouts to scalar fitness and structured feedback metrics. Fitness selects accepted candidates and the next elite, while feedback metrics condition policy proposals generated by memory-guided hill-climbing, macro-mutation, and crossover. We evaluate \textsc{MEMENTO} on two long-horizon embodied domains: Robosuite Franka Tower-of-Hanoi manipulation and AI2-THOR household interaction. \textsc{MEMENTO} outperforms Eureka and REvolve, adapted as code-as-policy evolutionary baselines, in task success and generalization to held-out Robosuite object configurations and unseen AI2-THOR scenes. Ablations show that zero-shot generation and unevolved evaluators fail to solve either domain, and that removing policy-search branches reduces performance. Finally, we deploy the best-evolved Robosuite policy on a physical Franka robot, demonstrating the feasibility of sim-to-real transfer of the evolved code-as-policy. Code, prompts, and videos are available at:
\url{https://github.com/sygkounas/MEMENTO}.

\end{abstract}





\section{Introduction}

Long-horizon embodied control tasks require policies to complete many dependent actions before task success can be evaluated \citep{mees2022calvin,shridhar2020alfred,srivastava2022behavior}. Earlier actions determine the task state encountered by later actions \citep{shridhar2020alfred,srivastava2022behavior}. Correct action order is therefore insufficient; each action must also be physically executed from the observed task state. This creates a long-horizon credit-assignment problem because task success is observed only after many dependent actions \citep{arjona2019rudder}.
Existing approaches reduce the effect of delayed feedback by adding external structure to policy learning or by changing how experience is generated and credited. Demonstration-based and hierarchical methods can constrain long-horizon behavior through trajectories, subgoal orderings, or reusable skills, while reinforcement-learning methods make delayed task feedback more usable through exploration, relabeling, temporal abstraction, or return redistribution \citep{zhao2023learning,wang2023mimicplay,shridhar2020alfred,padmakumar2022teach,andreas2017policy,andrychowicz2017her,pathak2017curiosity,ecoffet2021goexplore,sutton1999between,arjona2019rudder}. These approaches improve long-horizon learning, but they do not directly expose the decision logic of a failed policy or identify which part of that logic should be revised.

When policies are represented as executable programs, decision logic is written directly as program code and can thereby be inspected and revised. Large language models (LLMs) generate robot policy code that processes perception outputs and parameterizes control-primitive APIs, a formulation termed code-as-policy \citep{liang2022code}. This formulation has been extended to contact-rich control and video-conditioned policy generation \citep{genchip2024,xie2025robopro}. In long-horizon tasks, however, a generated program can execute without errors while its control logic fails to achieve the task \citep{rlgpt}, and failure feedback from rollouts has, therefore, been used to characterize policy errors and guide refinement \citep{roboinspector2025}. Code-as-policy improvement can then be formulated as execution-guided program search, in which candidate policy programs are executed in the task environment, evaluated based on observed behavior, and revised in response to failure feedback. Binary task success alone, however, is too sparse to guide the revision of a long-horizon policy program, and structured rollout feedback indicating which policy decisions to revise is difficult to specify by hand.

In genetic programming, candidate programs are executed and assigned a fitness score based on performance, which is then used to select programs for variation by micro-mutations that make small local changes, macro-mutations that modify larger program components, and crossover \citep{koza1992genetic,brameier2007linear,angeline1997subtree}. High-performing programs often require many candidate evaluations \citep{echevarrieta2024speeding}, making undirected syntactic variation costly in rollout-evaluated embodied domains. Language models have since been used as variation operators over executable programs, replacing syntactic program modifications with prompt-conditioned code proposals \citep{lehman2022elm,hemberg2024evolving}. LLM-driven code evolution has also been applied to executable policy programs, where candidate controllers are evaluated in task environments and selected for further revision \citep{hu2026mles}. In embodied control, LLM-guided evolutionary methods have evolved reward code via LLM-generated variation, with candidate reward functions selected based on policy performance \citep{ma2024eureka,hazra2025revolve}. However, existing code-as-policy and reward-code evolution methods primarily select among independently generated variants; they neither refine an accepted candidate through a local-search chain nor retain a memory of revisions that have already failed. Memetic algorithms support this form of refinement by combining evolutionary variation with local search on selected candidates \citep{moscato1989memetic,neri2012memetic}. The local-search component enables exploitation of an accepted candidate via iterative program revision, while mutation and crossover introduce additional variation via larger changes and recombination.

\begin{figure}[ht]
\centering
\includegraphics[width=\textwidth]{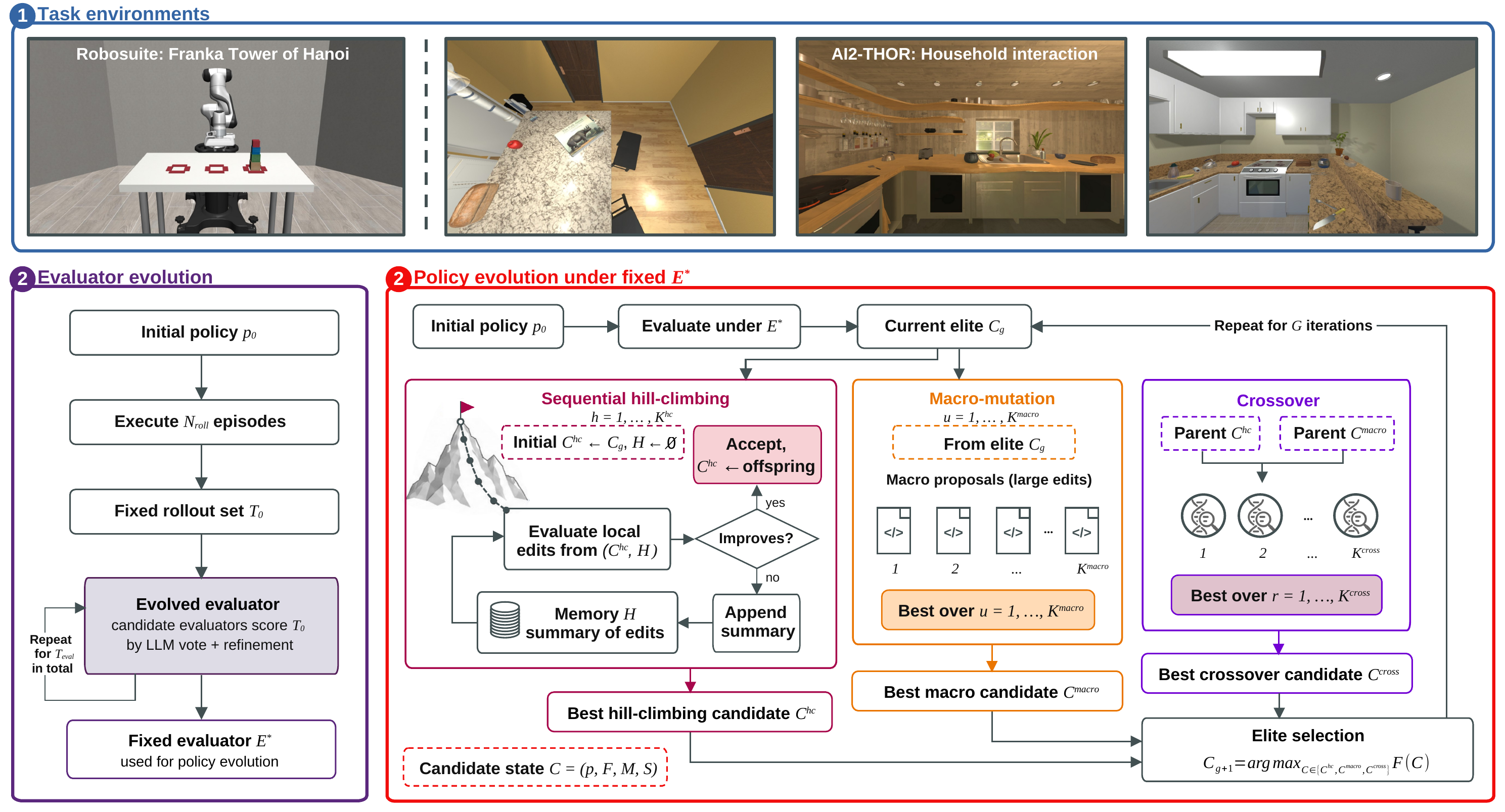}
\caption{ 
Conceptual overview of MEMENTO. The top row shows the task environments used in the study, including a Robosuite Franka Tower-of-Hanoi task and AI2-THOR household interaction scenes. The left panel illustrates evaluator evolution: candidate evaluators are generated, judged by an LLM majority vote, and iteratively refined to obtain the evolved evaluator $E^\star$. The right panel illustrates policy evolution under the fixed evaluator $E^\star$. An initial policy $p_0$ is evaluated to form the current elite candidate state $\mathcal{C}_g$, after which each generation applies three proposal mechanisms: sequential hill-climbing from the current accepted candidate with
rejected-proposal memory $\mathcal{H}$, macro-mutation from the elite,
and crossover between the best hill-climb and macro-mutation candidates. The best resulting candidate is selected as the next elite $\mathcal{C}_{g+1}$, and the process repeats for $G$ generations.
}
\label{fig:memento_overview}
\end{figure}

Local iterative improvement, search memory, and evolutionary variation via micro-mutation, macro-mutation, and crossover have not been combined in embodied policy programs that have been evaluated through environment rollouts. This limitation is addressed by \textsc{MEMENTO}, a memory-guided single-elite memetic search method for code-as-policy evolution in long-horizon embodied tasks. As illustrated in Figure~\ref{fig:memento_overview}, \textsc{MEMENTO} first evolves an evaluator \(E^\star\) that maps policy rollouts to scalar fitness and structured feedback metrics and then freezes it for code-as-policy search. Starting from an executable policy program, candidates are generated in each generation through three revision branches: sequential hill-climbing through micro-mutations from the current accepted candidate with rejected-proposal memory, macro-mutation from the elite, and crossover between the best hill-climb and macro-mutation candidates. The best branch output becomes the elite for the next generation. \textsc{MEMENTO} is evaluated on two long-horizon embodied domains: fully observed Robosuite Franka Tower-of-Hanoi \citep{zhu2020robosuite,havur2013case,duggan2026price}, in which object placements define the legal Hanoi state for subsequent pick-and-place moves, and partially observed AI2-THOR household interaction \citep{kolve2017ai2thor}, in which valid object interactions depend on egocentric visibility and scene-specific object identifiers. The final Robosuite Tower-of-Hanoi policy is also deployed on a physical Franka robot, showcasing the feasibility of transferring policies evolved through simulation to real-world settings. In particular, the contributions are as follows:

\begin{itemize}
    \item  a single-elite memetic search framework for executable policy programs that combines sequential local improvement with rejected-proposal memory, macro-mutation, and crossover;
    
    \item  an evaluator-evolution stage that produces a fixed evaluator before policy search, providing scalar fitness for selection and structured rollout feedback metrics for conditioning code-as-policy revisions;
    
    
    \item an empirical evaluation on two long-horizon embodied domains, including zero-shot, evaluator, and operator ablations, generalization across object-geometry and scene variations;
    

    \item a physical Franka deployment demonstrating sim-to-real transfer of the evolved Robosuite controller.

\end{itemize}

\section{Related Work}
\label{sec:related_work}

\paragraph{Long-horizon embodied control.}
Long-horizon embodied tasks have been approached by constraining policy optimization with prior information. Extended manipulation behavior has been learned from teleoperated or language-annotated demonstrations in ACT \citep{zhao2023learning}, Diffusion Policy \citep{chi2025diffusion}, MimicPlay \citep{wang2023mimicplay}, and BLADE \citep{liu2025blade}, and demonstration and interaction data for compositional tasks are provided by the ALFRED \citep{shridhar2020alfred}, TEACh \citep{padmakumar2022teach}, CALVIN \citep{mees2022calvin}, and BEHAVIOR \citep{srivastava2022behavior} benchmarks. Task structure has also been specified through subtask sequences or program decompositions, as in Policy Sketches \citep{andreas2017policy} and Neural Task Programming \citep{xu2018neural}, and ordering-constrained manipulation has been studied in rearrangement and Tower-of-Hanoi settings \citep{duggan2026price,kim2026humanoid}. Sparse-reward reinforcement learning instead addresses delayed task feedback through goal relabeling, exploration, temporal abstraction, return decomposition, and model-based rollouts \citep{andrychowicz2017her,pathak2017curiosity,ecoffet2021goexplore,sutton1999between,arjona2019rudder,hafner2023dreamerv3}. Across these approaches, delayed feedback is addressed by adding supervision, imposing temporal structure, or changing how experience is generated and credited during learning.

\paragraph{Language models for robot programs.}
Language models have been used to map task context to robot behavior, either by selecting among predefined skills or by generating executable programs. Action selection has been grounded with learned affordance and value estimates in SayCan \citep{ichter2023saycan}, and execution feedback and language-conditioned spatial reasoning have been incorporated into language-model planning in Inner Monologue \citep{huang2023innermonologue}, VoxPoser \citep{huang2023voxposer}, and LLM-Planner \citep{song2023llmplanner}. Program-structured task plans have been produced from program-like prompts in ProgPrompt \citep{singh2023progprompt}. Policy programs that process perception outputs and parameterize control-primitive APIs have been generated in Code as Policies \citep{liang2022code}, and in RoboPro \citep{xie2025robopro} and GenCHiP \citep{genchip2024} for visually grounded and contact-rich control. Rollout failures in generated policy programs have been characterized and used for failure-feedback refinement in RoboInspector \citep{roboinspector2025}, and execution summaries have been used to explain failed robot trials and propose corrective actions in REFLECT \citep{liu2023reflect}. Together, these works show how language models can produce, ground, and refine robot programs from task context and execution feedback.

\paragraph{LLM-driven evolutionary search.}
This line of research follows genetic programming, in which executable programs are varied through mutation and crossover and selected for fitness \citep{koza1992genetic}. Language models have since been used as the variation operators, with model-conditioned edits replacing syntactic operators in ELM \citep{lehman2022elm} and LLM\_GP \citep{hemberg2024evolving}. 
The methods that follow differ primarily in the object being searched. Discrete prompts have been evolved through model-generated mutation and crossover in EvoPrompt \citep{guo2023evoprompt} and self-referentially in Promptbreeder \citep{fernando2023promptbreeder}. Executable programs have been evolved by pairing a frozen model with a fixed evaluator in FunSearch \citep{romera2024mathematical} and, at codebase scale, in AlphaEvolve \citep{novikov2025alphaevolve}, and heuristics have been evolved through accumulated reflective feedback in ReEvo \citep{ye2024reevo} and through memetic local search in EoH-S \citep{liu2026eohs}. Reward code has been evolved from model-generated variation selected by policy performance in Eureka \citep{ma2024eureka}, and with population-based crossover and human-feedback fitness in REvolve \citep{hazra2025revolve}. REvolve has also been extended from reward-function discovery to the discovery of executable reinforcement-learning algorithms, in which candidate update rules are evaluated through full training runs \citep{sygkounas2026evolutionary}, while policy and environment programs are co-evolved in COvolve \citep{sygkounas2026covolve}. Programmatic control policies have been evolved directly in MLES \citep{hu2026mles}, where multimodal models generate candidate policy programs, selection uses quantitative task metrics, and new proposals are conditioned on behavioral evidence from rollouts. In these methods, the evaluator is fixed or externally specified, and the search object is a prompt, a heuristic, a reward, a program, or a program pair.

\paragraph{Memetic and memory-based search.}
Memetic algorithms combine population-level variation with local improvement of selected candidates \citep{moscato1989memetic,neri2012memetic}. Recombination has been coupled with local refinement in crossover hill-climbing \citep{lozano2004real}, and local search has been concentrated on promising candidates through local-search chains \citep{molina2010memetic}. Memory-based metaheuristics such as tabu search use search history to restrict recently explored moves and bias local improvement toward unexplored neighborhoods \citep{glover1997tabu}. \textsc{MEMENTO} adapts this structure to model-generated code-as-policy: a single accepted candidate is refined by a memory-guided local-search branch that records rejected proposals, along with macro-mutation and crossover, all of which are selected by the evolved rollout evaluator.

\section{Methodology}
\label{sec:methodology}
Sequential decision problems are modeled as finite-horizon POMDPs \(\mathcal{M}_{\mathrm{PO}}=(\mathcal{S},\mathcal{A},\mathcal{O},T,Z,R,\rho_0,L)\), where \(\mathcal{S}\) is the state space, \(\mathcal{A}\) the action space, \(\mathcal{O}\) the observation space, \(T:\mathcal{S}\times\mathcal{A}\to\Delta(\mathcal{S})\) the transition function, \(Z:\mathcal{S}\times\mathcal{A}\to\Delta(\mathcal{O})\) the observation function, \(R:\mathcal{S}\times\mathcal{A}\to\mathbb{R}\) the environment reward, \(\rho_0\) the initial-state distribution, and \(L\) the episode horizon \citep{kaelbling1998planning,puterman1994markov}. At time \(t=0,\ldots,L-1\), the environment has state \(s_t\in\mathcal{S}\), the agent receives observation \(o_t\in\mathcal{O}\), selects action \(a_t\in\mathcal{A}\), and the environment transitions according to \(T\). Task completion is represented by an episode-level success indicator \(\sigma(\tau)\in\{0,1\}\), defined on an executed trajectory \(\tau\).

The method does not directly optimize the environment reward \(R\). Instead, it searches over executable code-as-policy programs. Each candidate program \(p\in\mathcal{P}\) defines a deterministic stateful policy with internal program-state space \(\mathcal{Z}_p\). At the start of each episode, the program initializes its internal state \(z_0\in \mathcal{Z}_p\). At timestep \(t\), the executable program maps the current observation and internal program state to an action and an updated program state,
\[
(a_t,z_{t+1}) = p(o_t,z_t),
\qquad
o_t\in\mathcal{O},\; a_t\in\mathcal{A}.
\]
The induced policy is denoted by \(\pi_p\). Executing \(\pi_p\) in \(\mathcal{M}_{\mathrm{PO}}\) for episode \(e\) induces a trajectory
\(\tau_e(p)\). For a rollout budget of \(N_{\mathrm{roll}}\) episodes, let
\[
\mathcal{T}_{N_{\mathrm{roll}}}(p)
=
\{\tau_1(p),\ldots,\tau_{N_{\mathrm{roll}}}(p)\}
\]
denote the resulting rollout set. A fixed evaluator \(E\) maps this rollout set to program-level fitness and feedback metrics,
\[
E\!\left(\mathcal{T}_{N_{\mathrm{roll}}}(p)\right)
=
\bigl(F(p),\mathbf{M}(p)\bigr).
\]
Here \(F(p)\in[0,1]\) is the aggregate evaluator fitness over the \(N_{\mathrm{roll}}\) episodes, and \(\mathbf{M}(p)\) is the collection of rollout-level feedback metrics used to condition later code-as-policy revisions. The empirical success rate is computed from the task success indicator,
\[
S(p)
=
\frac{1}{N_{\mathrm{roll}}}
\sum_{e=1}^{N_{\mathrm{roll}}}
\sigma\!\left(\tau_e(p)\right).
\]
Evaluating a program defines the candidate state \(\mathcal{C}(p)=(p,F(p),\mathbf{M}(p),S(p))\), which stores the program, the evaluator outputs, and the task-derived success rate. For any candidate state \(\mathcal{C}\), \(F(\mathcal{C})\) denotes the stored fitness value in \(\mathcal{C}\).

\subsection{Evaluator Evolution}
\label{subsec:evaluator_evolution}

Evaluator evolution searches for evaluator programs that map rollout data to scalar fitness and feedback metrics for code-as-policy search. The initial policy \(p_0\) is executed for \(N_{\mathrm{roll}}\) episodes, yielding the fixed rollout set \(\mathcal{T}_0=\mathcal{T}_{N_{\mathrm{roll}}}(p_0)\) and empirical success rate \(S_0=S(p_0)\). The same rollout set is used for all evaluator candidates during evaluator evolution. At evaluator generation \(i\), the search maintains a pool \(\mathcal{E}_i=\{E_i^{(1)},\ldots,E_i^{(n)}\}\) of \(n\) evaluator programs. The initial pool \(\mathcal{E}_0\) contains \(n\) LLM-generated evaluator programs. For \(i>0\), \(\mathcal{E}_i\) is obtained by applying the evaluator macro-mutation operator \(q_{\mathrm{eval}}^{\mathrm{macro}}\) to the selected evaluator from the previous generation, \(E_{i-1}^\star\).

Each evaluator candidate is applied to the shared rollout set and produces candidate-specific program-level outputs,
\[
E_i^{(j)}(\mathcal{T}_0)
=
\bigl(F_i^{(j)}(p_0),\mathbf{M}_i^{(j)}(p_0)\bigr).
\]
The empirical success rate \(S_0\) is computed from the task success indicator and is shared across the evaluated pool. Let
\[
\mathcal{D}_i
=
\left\{
\bigl(E_i^{(j)},F_i^{(j)}(p_0),\mathbf{M}_i^{(j)}(p_0),S_0\bigr)
\right\}_{j=1}^{n}
\]
denote the scored evaluator pool. The selected evaluator is chosen by majority voting over \(k\) independent LLM judgments of \(\mathcal{D}_i\),
\[
E_i^\star = \mathcal{V}_k(\mathcal{D}_i).
\]
The voting step uses LLM-as-judge evaluation to compare the scored evaluator pool, aggregating by majority over \(k\) independent judgments \citep{zheng2023judging,wang2023selfconsistency}. After \(T_{\mathrm{eval}}\) evaluator generations, \(E^\star=E_{T_{\mathrm{eval}}-1}^\star\) is fixed and used for code-as-policy evolution. The full evaluator evolution process is outlined in Algorithm~\ref{alg:eval_evolution}.  

\begin{algorithm}[H]
\caption{Evaluator Evolution}
\label{alg:eval_evolution}
\begin{algorithmic}[1]
\Statex \textbf{Require:} initial policy \(p_0\), pool size \(n\), rollout budget \(N_{\mathrm{roll}}\), voting budget \(k\), evaluator generations \(T_{\mathrm{eval}}\)

\State Execute \(p_0\) for \(N_{\mathrm{roll}}\) episodes; store \(\mathcal{T}_0=\mathcal{T}_{N_{\mathrm{roll}}}(p_0)\) and \(S_0=S(p_0)\)
\State Generate the initial evaluator pool \(\{E_0^{(1)},\ldots,E_0^{(n)}\}\)

\For{\(i=0\) to \(T_{\mathrm{eval}}-1\)}
\State \(\mathcal{D}_i \gets \emptyset\)
\For{\(j=1\) to \(n\)}
    \State \((F_i^{(j)},\mathbf{M}_i^{(j)}) \gets E_i^{(j)}(\mathcal{T}_0)\)
    \State Add \((E_i^{(j)},F_i^{(j)},\mathbf{M}_i^{(j)},S_0)\) to \(\mathcal{D}_i\)
\EndFor
\State \(E_i^\star \gets \mathcal{V}_k(\mathcal{D}_i)\)
\Comment{majority vote over \(k\) independent LLM judgments}
\State \(\{E_{i+1}^{(j)}\}_{j=1}^{n} \leftarrow \{q_{\mathrm{eval}}^{\mathrm{macro}}(E_i^\star)\}_{j=1}^{n}\)
\Comment{unused when $i=T_{\mathrm{eval}}-1$}
\EndFor

\State \Return \(E^\star \gets E_{T_{\mathrm{eval}}-1}^\star\)
\end{algorithmic}
\end{algorithm}

\subsection{Policy Evolution Under the Fixed Evaluator}
\label{subsec:policy_evolution}

After \(E^\star\) is fixed, policy search proceeds from a single elite candidate state \(\mathcal{C}_g=\mathcal{C}(p_g)\). Each generation explores three branches: a hill-climb branch of sequential micro-mutations, a macro-mutation branch, and a crossover branch (cf. Figure~\ref{fig:memento_overview}). Let \(q_{\mathrm{hc}}\), \(q_{\mathrm{macro}}\), and \(q_{\mathrm{cross}}\) denote LLM-conditioned proposal distributions that take evaluated candidate states as input and return new policy programs. Proposals are conditioned on the program text and the evaluator outputs \(F\) and \(\mathbf{M}\) of the input candidate states; the success rate \(S\) is recorded for reporting and is not used during search. The complete policy evolution procedure is presented in Algorithm~\ref{alg:policy_evolution}. 

\begin{algorithm}[ht]
\caption{Single-Elite Memetic Code-as-Policy Evolution}
\label{alg:policy_evolution}
\footnotesize
\begin{algorithmic}[1]
\Require Initial policy $p_0$; evaluator $E^\star$; generations $G$;
budgets $N_{\mathrm{roll}}, K_{\mathrm{hc}}, K_{\mathrm{macro}}, K_{\mathrm{cross}}$
\State $\mathcal{C}_0 \gets \mathcal{C}(p_0)$
\Comment{evaluate initial policy under $E^\star$}
\For{$g = 0$ to $G-1$}
\State $\mathcal{C}^{\mathrm{hc}} \gets \mathcal{C}_g$,
$\mathcal{H} \gets \emptyset$
\Comment{hill-climb branch}
\For{$h = 1$ to $K_{\mathrm{hc}}$}
    \State $p \sim q_{\mathrm{hc}}(\mathcal{C}^{\mathrm{hc}}, \mathcal{H})$
    \State $\mathcal{C}^{\mathrm{new}} \gets \mathcal{C}(p)$
    \Comment{execute $\pi_p$ for $N_{\mathrm{roll}}$ episodes, evaluate under $E^\star$}
    \If{$F(\mathcal{C}^{\mathrm{new}}) \geq F(\mathcal{C}^{\mathrm{hc}})$}
        \State $\mathcal{C}^{\mathrm{hc}} \gets \mathcal{C}^{\mathrm{new}}$
    \Else
        \State $\mathcal{H} \gets \operatorname{append}(\mathcal{H}, \mathrm{history}(p))$
    \EndIf
\EndFor
\For{$u = 1$ to $K_{\mathrm{macro}}$}
    \Comment{macro-mutation branch}
    \State $p_u \sim q_{\mathrm{macro}}(\mathcal{C}_g)$;
    $\mathcal{C}_u^{\mathrm{macro}} \gets \mathcal{C}(p_u)$
\EndFor
\State $\mathcal{C}^{\mathrm{macro}} \gets
\operatorname*{arg\,max}_{\mathcal{C}\in\{\mathcal{C}_g,
\mathcal{C}_1^{\mathrm{macro}},\ldots,
\mathcal{C}_{K_{\mathrm{macro}}}^{\mathrm{macro}}\}}
F(\mathcal{C})$
\For{$r = 1$ to $K_{\mathrm{cross}}$}
    \Comment{crossover branch}
    \State $p_r \sim q_{\mathrm{cross}}(\mathcal{C}^{\mathrm{hc}},
    \mathcal{C}^{\mathrm{macro}})$;
    $\mathcal{C}_r^{\mathrm{cross}} \gets \mathcal{C}(p_r)$
\EndFor
\State $\mathcal{C}^{\mathrm{cross}} \gets
\operatorname*{arg\,max}_{\mathcal{C}\in\{\mathcal{C}_1^{\mathrm{cross}},
\ldots,\mathcal{C}_{K_{\mathrm{cross}}}^{\mathrm{cross}}\}}
F(\mathcal{C})$
\State $\mathcal{C}_{g+1} \gets
\operatorname*{arg\,max}_{\mathcal{C}\in\{\mathcal{C}^{\mathrm{hc}},
\mathcal{C}^{\mathrm{macro}},\mathcal{C}^{\mathrm{cross}}\}}
F(\mathcal{C})$
\Comment{elitist selection}
\EndFor
\State \Return $\mathcal{C}_G$
\end{algorithmic}
\end{algorithm}

\paragraph{Hill-climb branch.}
For any proposed policy program \(p\), let \(\mathrm{history}(p)\) denote the edit summary embedded in the program text, describing the change made relative to its parent. The hill-climb branch maintains a current accepted candidate state \(\mathcal{C}_{g,h}^{\mathrm{hc}}=\mathcal{C}(p_{g,h}^{\mathrm{hc}})\) and a rejected-proposal memory \(\mathcal{H}_{g,h}\), where \(\mathcal{H}_{g,h}\) is an ordered list of edit summaries from rejected hill-climb proposals in generation \(g\). Edit summaries of accepted proposals remain embedded in the accepted program text and are therefore visible to subsequent proposals through \(\mathcal{C}_{g,h}^{\mathrm{hc}}\); \(\mathcal{H}_{g,h}\) records only rejected edits. The branch is initialized from the generation elite, with \(\mathcal{C}_{g,0}^{\mathrm{hc}}=\mathcal{C}_g\), and starts with an empty rejected-proposal memory, \(\mathcal{H}_{g,0}=\emptyset\). At step \(h=1,\ldots,K_{\mathrm{hc}}\), a new policy program is sampled from the current accepted candidate state and the rejected-proposal memory:
\[
p_{g,h}^{\mathrm{new}}
\sim
q_{\mathrm{hc}}\!\left(
\mathcal{C}_{g,h-1}^{\mathrm{hc}},
\mathcal{H}_{g,h-1}
\right).
\]
The proposed program is executed and evaluated under \(E^\star\), producing
\[
\mathcal{C}_{g,h}^{\mathrm{new}}
=
\mathcal{C}(p_{g,h}^{\mathrm{new}}).
\]
The proposal is accepted if its evaluated fitness is at least that of the current accepted program:
\begin{equation}
\mathcal{C}_{g,h}^{\mathrm{hc}}
=
\begin{cases}
\mathcal{C}_{g,h}^{\mathrm{new}},
& \text{if }
F(\mathcal{C}_{g,h}^{\mathrm{new}})
\geq
F(\mathcal{C}_{g,h-1}^{\mathrm{hc}}),\\[2pt]
\mathcal{C}_{g,h-1}^{\mathrm{hc}},
& \text{otherwise.}
\end{cases}
\end{equation}
If the proposal is accepted, the rejected-proposal memory is unchanged: \(\mathcal{H}_{g,h}=\mathcal{H}_{g,h-1}\). If the proposal is rejected, its edit summary is appended to the memory: \(\mathcal{H}_{g,h}=\operatorname{append}(\mathcal{H}_{g,h-1}, \mathrm{history}(p_{g,h}^{\mathrm{new}}))\). The branch output is the accepted candidate state after the final hill-climb step,
\[
\mathcal{C}_{g}^{\mathrm{hc}}
=
\mathcal{C}_{g,K_{\mathrm{hc}}}^{\mathrm{hc}} .
\]

\paragraph{Macro-mutation branch.}
For \(u=1,\ldots,K_{\mathrm{macro}}\), an independent mutation of the generation elite is sampled and evaluated,
\[
p_{g,u}^{\mathrm{macro}}
\sim
q_{\mathrm{macro}}(\mathcal{C}_g),
\qquad
\mathcal{C}_{g,u}^{\mathrm{macro}}
=
\mathcal{C}(p_{g,u}^{\mathrm{macro}}).
\]
The branch output is the candidate state with the highest fitness among the generation elite and the macro-mutation candidates:
\begin{equation}
\mathcal{C}_g^{\mathrm{macro}}
=
\operatorname*{arg\,max}_{\mathcal{C}\in
\{\mathcal{C}_g,\mathcal{C}_{g,1}^{\mathrm{macro}},\ldots,
\mathcal{C}_{g,K_{\mathrm{macro}}}^{\mathrm{macro}}\}}
F(\mathcal{C}) .
\end{equation}
\paragraph{Crossover branch.}
For \(r=1,\ldots,K_{\mathrm{cross}}\), a recombination of the two branch outputs is sampled and evaluated,
\[
p_{g,r}^{\mathrm{cross}}
\sim
q_{\mathrm{cross}}\!\left(\mathcal{C}_g^{\mathrm{hc}},
\mathcal{C}_g^{\mathrm{macro}}\right),
\qquad
\mathcal{C}_{g,r}^{\mathrm{cross}}
=
\mathcal{C}(p_{g,r}^{\mathrm{cross}}).
\]
The crossover proposal is conditioned on both selected candidate states, using their policy programs and evaluator outputs. The branch output is the evaluated crossover candidate with the highest program fitness:
\begin{equation}
\mathcal{C}_g^{\mathrm{cross}}
=
\operatorname*{arg\,max}_{\mathcal{C}\in
\{\mathcal{C}_{g,1}^{\mathrm{cross}},\ldots,
\mathcal{C}_{g,K_{\mathrm{cross}}}^{\mathrm{cross}}\}}
F(\mathcal{C}) .
\end{equation}

\paragraph{Generation elite selection.}
The branch output with the highest program fitness is selected as the next-generation elite and used as the parent for generation \(g+1\):
\begin{equation}
\mathcal{C}_{g+1}
=
\operatorname*{arg\,max}_{\mathcal{C}\in\{\mathcal{C}_g^{\mathrm{hc}},
\mathcal{C}_g^{\mathrm{macro}},\mathcal{C}_g^{\mathrm{cross}}\}}
F(\mathcal{C}) .
\end{equation}

\section{Experiments}
\label{sec:experiments}

\begin{figure}[ht]
\centering

\begin{minipage}{0.72\linewidth}
\centering
\includegraphics[
  width=\linewidth,
  trim=0 20pt 0 25pt,
  clip
]{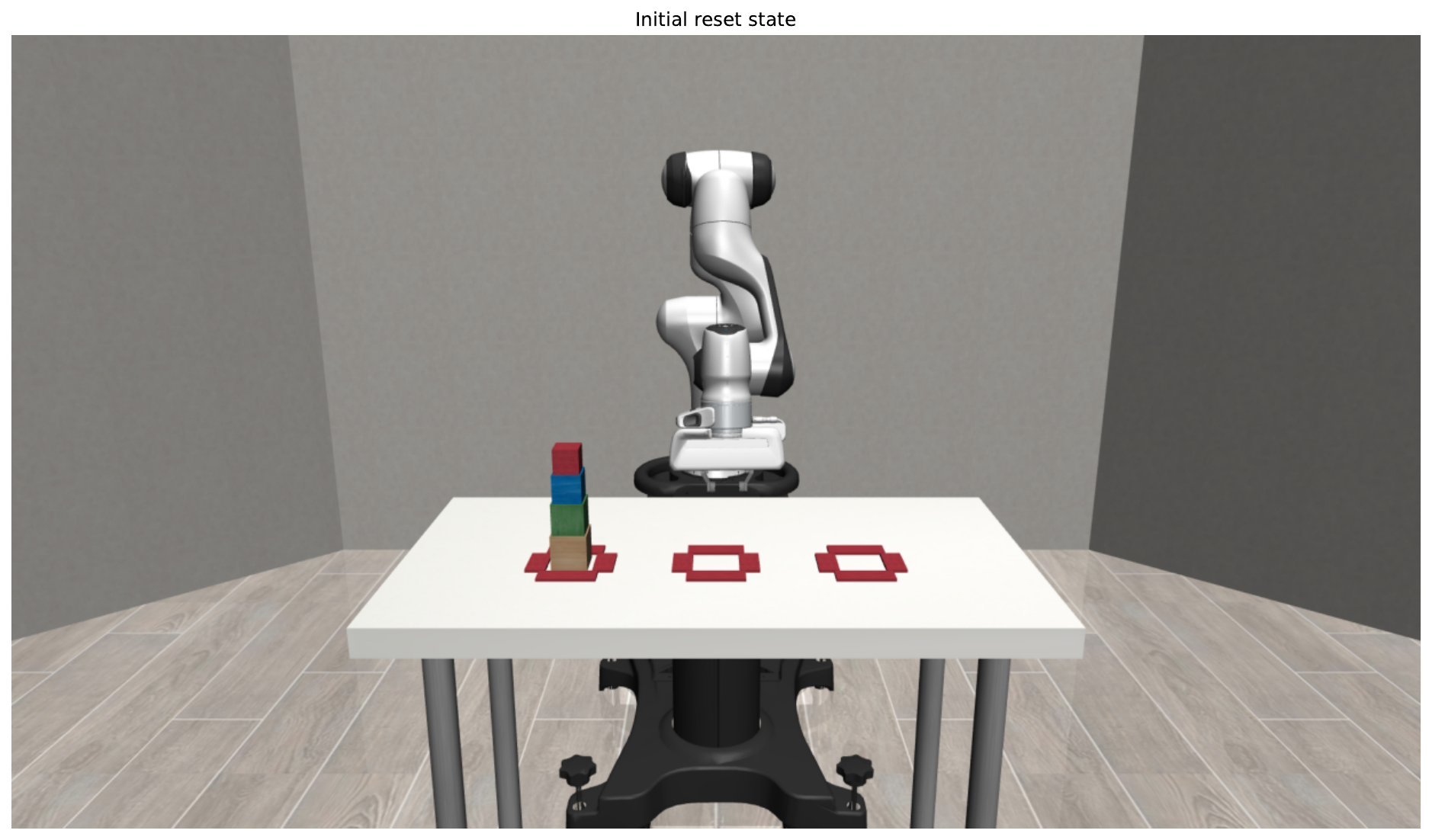}\\[-0.2em]
{\small (a) Robosuite Franka Tower-of-Hanoi}
\end{minipage}

\vspace{0.8em}

\begin{tabular}{ccc}
\begin{minipage}{0.31\linewidth}
  \centering
  \includegraphics[
    width=\linewidth,
    trim=0 20pt 0 25pt,
    clip
  ]{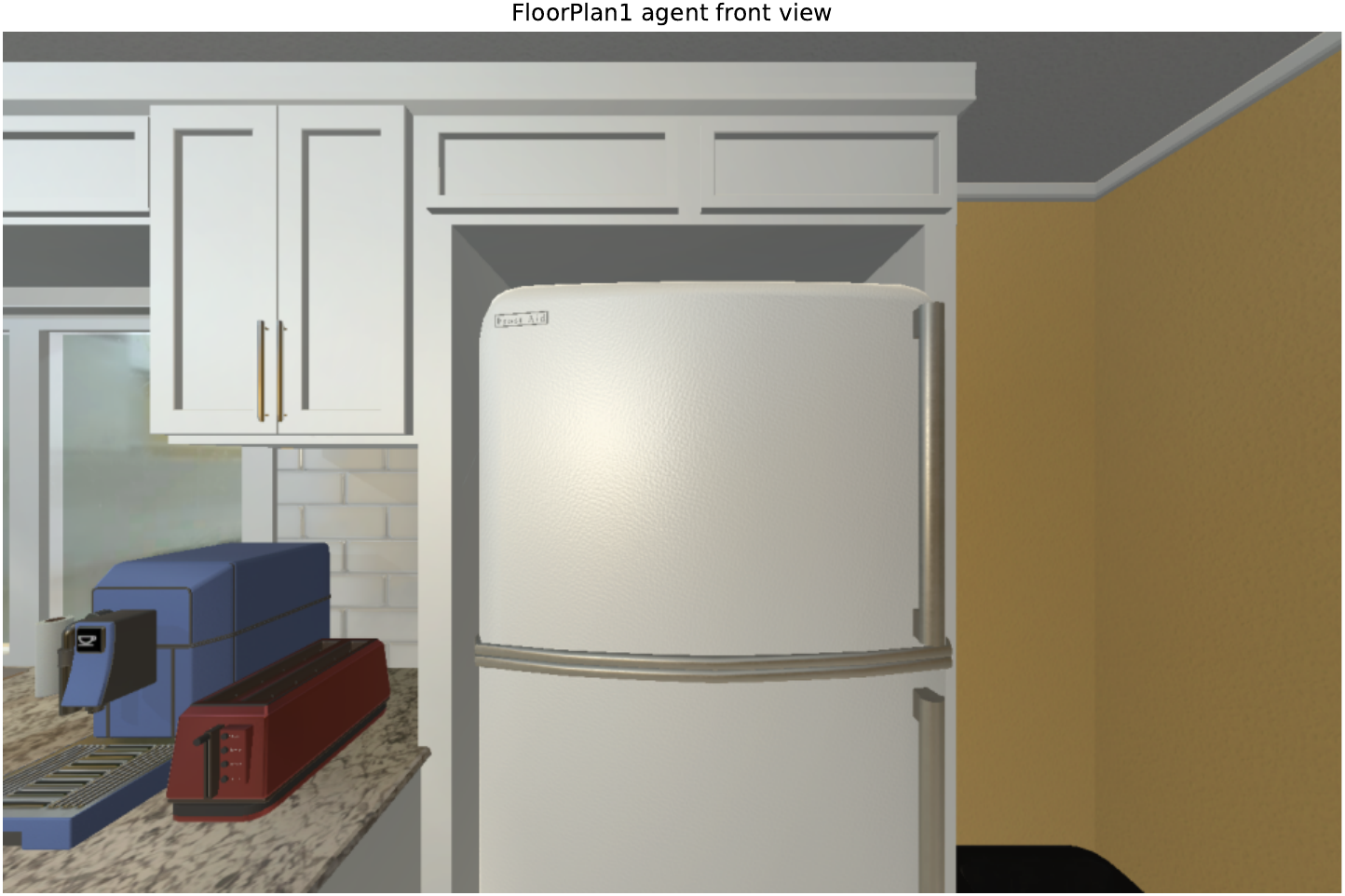}\\[-0.2em]
  {\small \texttt{FloorPlan1}}
\end{minipage}
&
\begin{minipage}{0.31\linewidth}
  \centering
  \includegraphics[
    width=\linewidth,
    trim=0 20pt 0 25pt,
    clip
  ]{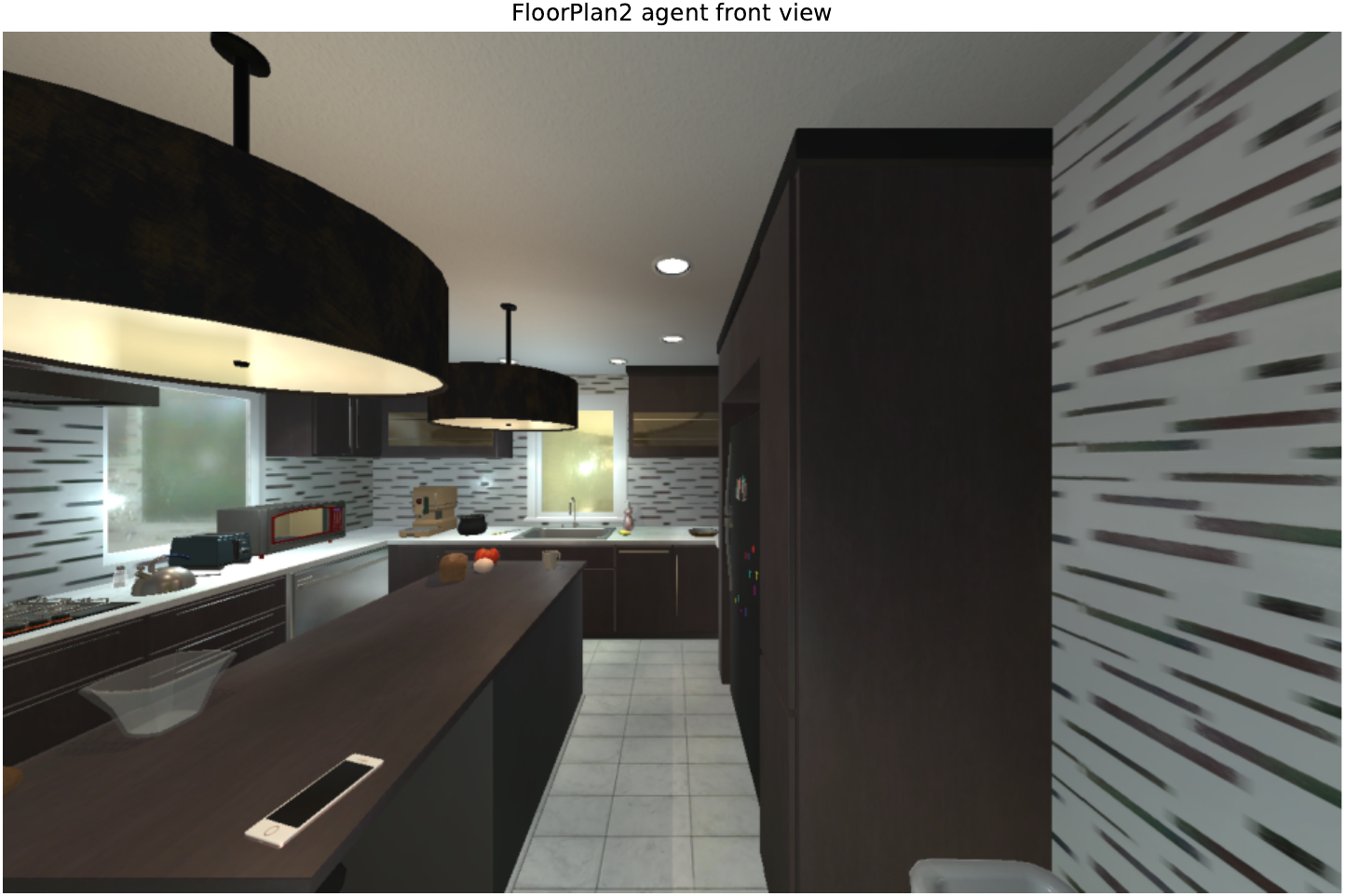}\\[-0.2em]
  {\small \texttt{FloorPlan2}}
\end{minipage}
&
\begin{minipage}{0.31\linewidth}
  \centering
  \includegraphics[
    width=\linewidth,
    trim=0 20pt 0 25pt,
    clip
  ]{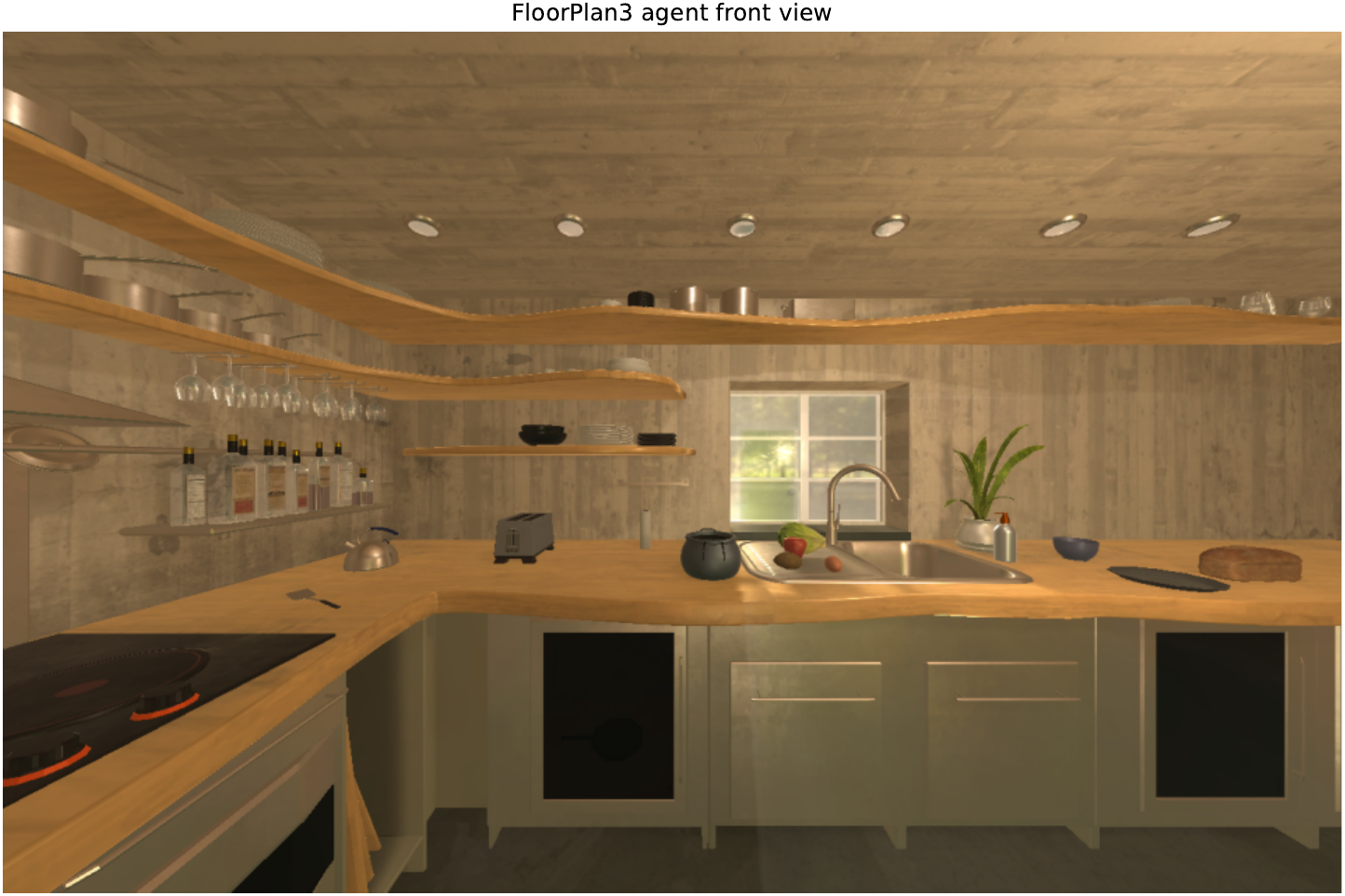}\\[-0.2em]
  {\small \texttt{FloorPlan3}}
\end{minipage}
\end{tabular}

\vspace{0.2em}
{\small (b) AI2-THOR training kitchens}

\caption{Training task environments. The Robosuite task is a four-object Tower-of-Hanoi manipulation problem in which the policy must move the full stack from the source container to the target container while preserving legal size order. The AI2-THOR task uses three training kitchens, \texttt{FloorPlan1}--\texttt{FloorPlan3}, which differ in layout, object placement, visibility, and receptacle configuration.}
\label{fig:task_envs}
\end{figure}

We evaluate \textsc{MEMENTO} on two long-horizon embodied domains with different sources of execution dependence. In Robosuite Franka Tower-of-Hanoi \citep{zhu2020robosuite,havur2013case,duggan2026price}, each pick-and-place operation changes the object configuration that determines the remaining legal moves. In AI2-THOR household interaction \citep{kolve2017ai2thor}, valid actions are constrained by egocentric visibility, scene-specific object identifiers, and object-interaction preconditions. The two domains, therefore, test code-as-policy search under structured manipulation state and partially observed household interaction. Figure~\ref{fig:task_envs} shows the training task settings used in the experiments.

\paragraph{Robosuite Franka Tower-of-Hanoi.}
The Robosuite domain is a four-cube Tower-of-Hanoi task with a Franka Panda arm and three containers. At reset, all cubes are stacked on container A in legal size order. Success requires moving the complete tower to container C while leaving containers A and B empty. An episode terminates on success, timeout, or an invalid Hanoi placement in which a larger cube is placed on a smaller cube. For \(n\) cubes, the shortest legal solution requires \(2^n-1\) transfers; for \(n=4\), this gives \(15\) ordered pick-and-place operations.

\paragraph{AI2-THOR household interaction.}
The AI2-THOR domain uses the task \texttt{apple\_in\_microwave\_on\_bread\_in\_fridge}. Success requires placing the Apple in the Microwave, toggling the Microwave on, and placing the Bread in the Fridge. The agent receives egocentric observations, and object actions are valid only when the corresponding scene-specific object identifier appears in the current action space. An episode terminates on success, policy-issued \texttt{done}, or the step limit. Policy evolution uses \texttt{FloorPlan1}--\texttt{FloorPlan3}; cross-scene transfer is evaluated on held-out kitchens \texttt{FloorPlan4}--\texttt{FloorPlan29}.

The full observation and action interfaces are reported in Appendix~\ref{app:franka_cap_obs_action} and Appendix~\ref{app:ai2thor_env}; Robosuite domain-randomization details are reported in Appendix~\ref{app:franka_heavy_dr}. Held-out Robosuite object variants are shown in Appendix~\ref{app:gen_hanoi}, and additional AI2-THOR scene examples are shown in Appendix~\ref{app:ai2thor_env}.

\paragraph{Training setup.} 
\textsc{MEMENTO} is run for \(G=5\) generations. Generation~0 denotes the evaluated initial policy $p_0$ and involves no policy search; each subsequent generation applies the three proposal branches. Each generation uses $K_{\mathrm{hc}}=10$ hill-climb proposals, $K_{\mathrm{macro}}=10$ macro-mutation proposals, and $K_{\mathrm{cross}}=4$ crossover proposals, giving a policy-candidate budget of $B=K_{\mathrm{hc}}+K_{\mathrm{macro}}+K_{\mathrm{cross}}=24$ new candidates per generation. Each candidate policy is evaluated for $N_{\mathrm{roll}}=10$ rollout episodes under the fixed evolved evaluator $E^\star$. Evaluator evolution uses a pool of $n=8$ evaluator candidates per generation for $T_{\mathrm{eval}}=5$ evaluator generations, with selection by majority vote over $k=10$ independent LLM judgments. Results are reported as mean~$\pm$~standard deviation across three random seeds. Curves report per-generation best-so-far fitness and empirical success rate, and tables report final-generation performance. Evaluator and code-as-policy proposals are generated with GPT-5.2 \citep{openai2025gpt52}.

\paragraph{Baselines.}
Eureka \citep{ma2024eureka} and REvolve \citep{hazra2025revolve} are adapted from reward-code evolution to code-as-policy evolution: the search object is the executable policy program, and candidate fitness is computed by the same fixed evolved evaluator $E^\star$ used by MEMENTO. All methods start from the same seed-specific initial policy, use the same language model and domain prompt templates, and receive the same budget of $B=24$ evaluated candidates per generation for $G=5$ generations. The adapted Eureka samples independent macro-mutations of the current best program in each generation and selects the highest-fitness candidate. The adapted REvolve maintains a population of policy programs with fitness-based
selection and LLM-generated crossover and macro-mutation, with $E^\star$ fitness replacing its human-feedback fitness.

\section{Results}
\label{sec:main_results}

\subsection{Main Evolution Results}
\label{subsec:main_evolution_results}

\begin{figure}[ht]
\centering

\begin{minipage}{\linewidth}
    \centering
    \includegraphics[width=0.9\linewidth]{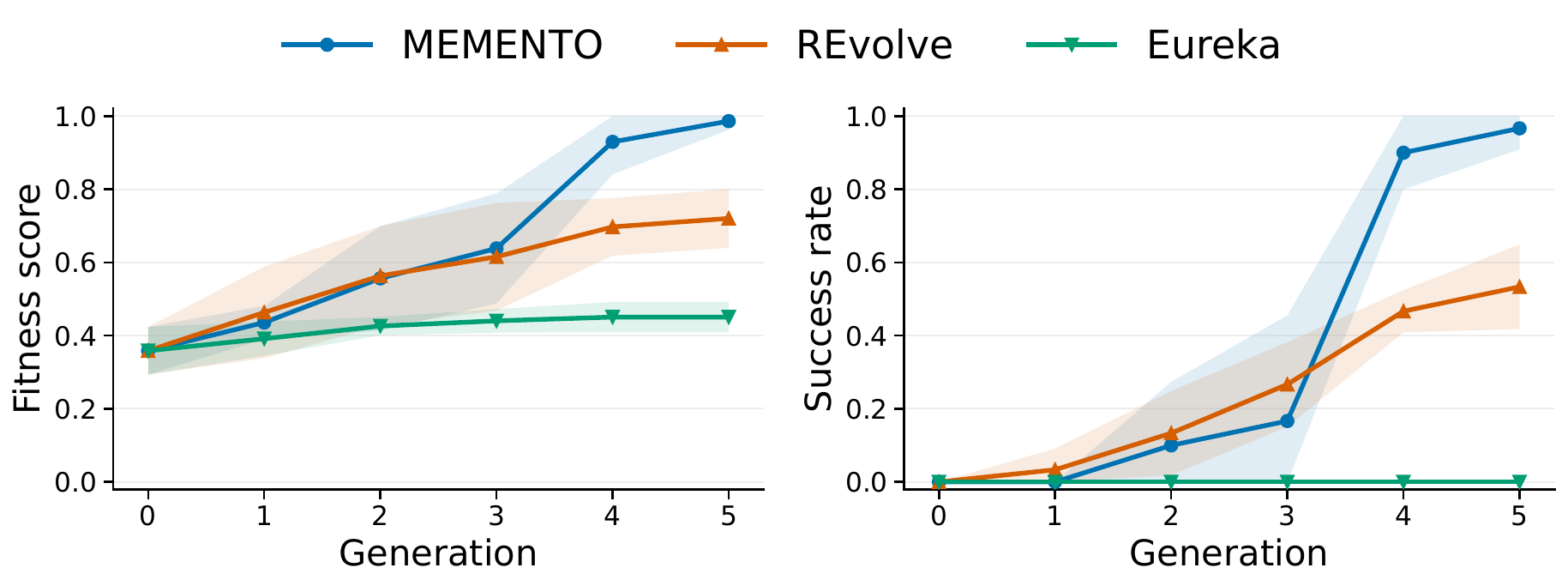}
    \vspace{-0.5em}
    \centerline{\small (a) Robosuite Tower-of-Hanoi}
\end{minipage}

\vspace{0.8em}

\begin{minipage}{\linewidth}
    \centering
    \includegraphics[width=0.9\linewidth]{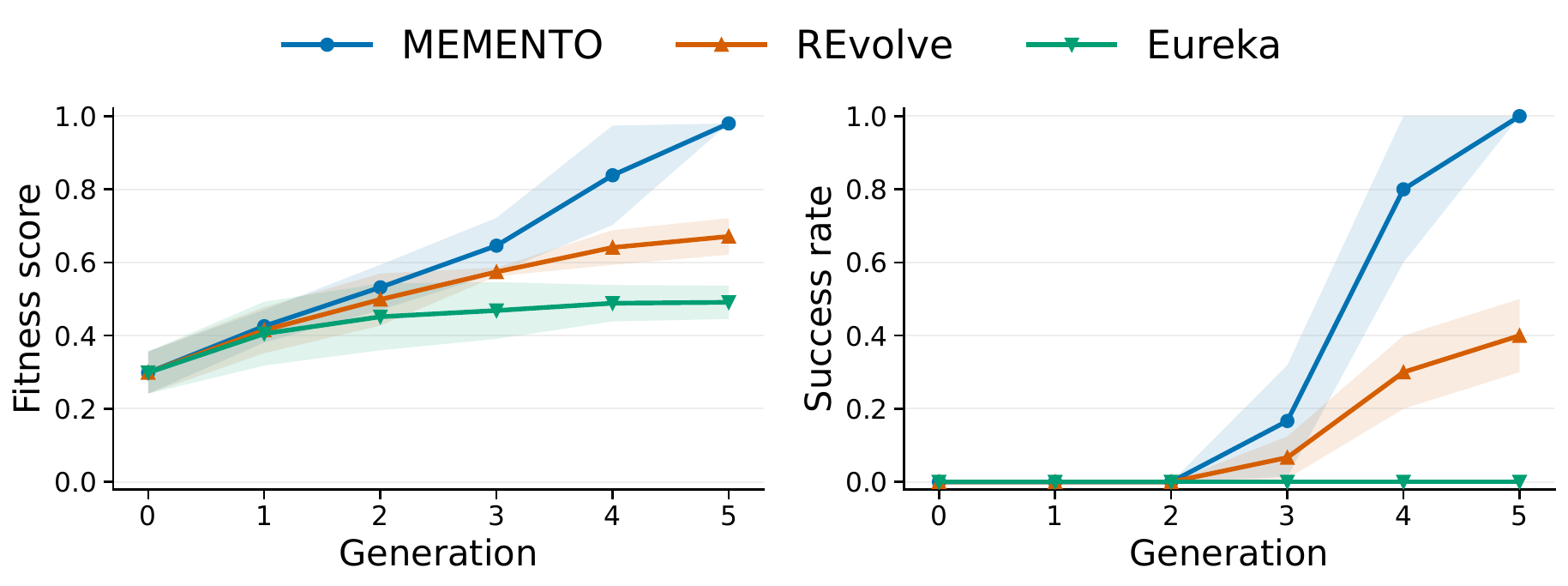}
    \vspace{-0.5em}
    \centerline{\small (b) AI2-THOR household interaction}
\end{minipage}

\caption{Main evolution results on Robosuite Tower-of-Hanoi and AI2-THOR household interaction. Subfigure~(a) reports Robosuite and Subfigure~(b) reports AI2-THOR. Within each subfigure, the left plot shows the best-so-far evaluator fitness, and the right plot shows the empirical success rate over generations. Lines denote the mean across three random seeds, and shaded regions denote one standard deviation for \textsc{MEMENTO}, REvolve, and Eureka.}
\label{fig:main_evolution}
\end{figure}

We first evaluate whether the proposed search structure improves code-as-policy programs more effectively than LLM-guided evolutionary baselines under the same rollout budget. Figure~\ref{fig:main_evolution} reports best-so-far evaluator fitness and empirical success rate across generations for \textsc{MEMENTO}, REvolve, and Eureka. Results are reported as mean \(\pm\) standard deviation across three random seeds, with all methods initialized from the same seed-specific policy at generation~\(0\).

By generation~5, \textsc{MEMENTO} achieves the strongest final performance in both domains. On Robosuite Tower-of-Hanoi, \textsc{MEMENTO} reaches \(\mathbf{0.99 \pm 0.02}\) fitness and \(\mathbf{0.97 \pm 0.06}\) success, compared with \(0.72 \pm 0.08\) fitness and \(0.53 \pm 0.12\) success for REvolve. Eureka reaches \(0.45 \pm 0.04\) fitness and produces no successful rollouts. On AI2-THOR, \textsc{MEMENTO} reaches \(\mathbf{0.98 \pm 0.00}\) fitness and \(\mathbf{1.00 \pm 0.00}\) success, compared with \(0.67 \pm 0.05\) fitness and \(0.40 \pm 0.10\) success for REvolve; Eureka again produces no successful rollouts. These results show that, under the same rollout evaluator and candidate budget, \textsc{MEMENTO}'s memory-guided local refinement, macro-mutation, and crossover produce task-completing code-as-policy programs more reliably than the mutation-only and population-style evolutionary baselines.

Because candidate selection is performed using the fixed evaluator fitness \(F\) returned by \(E^\star\), while empirical success \(S\) is computed independently from the task-success predicate, we examined the correspondence between these two quantities. For each domain, the per-generation selected candidates, i.e., the elite \(\mathcal{C}_g\) for \textsc{MEMENTO} and the reported best candidate for each baseline, provide \(54\) paired values \((F,S)\), obtained from three methods, three seeds, and six generations. Fitness and success are positively correlated in both domains: Robosuite Tower-of-Hanoi gives Pearson \(r=0.922\) and Spearman \(\rho=0.755\), while AI2-THOR gives Pearson \(r=0.871\) and Spearman \(\rho=0.801\). These correlations indicate that the fixed evaluator provides a selection signal aligned with task success: candidates with higher evaluator fitness generally also achieve higher empirical success.

\subsection{Final Performance and Generalization}
\label{subsec:final_generalization}

Table~\ref{tab:generalization} reports the success rate of the final selected policy after search on the original evaluation domains and on held-out generalization domains. For Robosuite Franka Tower-of-Hanoi, Cube denotes the cube-based environment used during policy evolution, while Asym. and Cyl. replace the manipulated cubes with held-out object geometries. These variants preserve the Hanoi objective, object identifiers, container layout, and structured observation interface, but change the grasp envelope and contact geometry of the manipulated objects; details and examples are given in Appendix~\ref{app:gen_hanoi}. For AI2-THOR, Search scenes denote the kitchens used during policy evolution, while Held-out scenes denote unseen kitchens with different layouts, object instances, visibility patterns, receptacle placements, and scene-specific object identifiers; environment details and scene examples are given in Appendix~\ref{app:ai2thor_env}. SAC+HER is included as a Robosuite-only interaction-learning baseline on the original cube task. Unlike the LLM-guided policy-code methods, SAC+HER is trained with goal-conditioned HER rewards rather than evaluator-conditioned code revision; its observation interface, training configuration, and sparse/dense reward definitions are reported in Appendices~\ref{app:franka_sac_obs_action}--\ref{app:franka_dense_sac_her}.

\begin{table*}[h]
\centering
\small
\renewcommand{\arraystretch}{1.08}
\setlength{\tabcolsep}{4.2pt}
\resizebox{1.00\textwidth}{!}{%
\begin{tabular}{lccccc}
\toprule
& \multicolumn{3}{c}{Robosuite Franka Tower-of-Hanoi}
& \multicolumn{2}{c}{AI2-THOR household interaction} \\
\cmidrule(lr){2-4}\cmidrule(lr){5-6}
Method & Cube & Asym. & Cyl. & Search scenes & Held-out scenes \\
\midrule
Eureka
& \(0.00 \pm 0.00\)
& \(0.00 \pm 0.00\)
& \(0.00 \pm 0.00\)
& \(0.00 \pm 0.00\)
& \(0.00 \pm 0.00\) \\

REvolve
& \(0.53 \pm 0.12\)
& \(0.23 \pm 0.15\)
& \(0.00 \pm 0.00\)
& \(0.40 \pm 0.10\)
& \(0.08 \pm 0.04\) \\

\textsc{MEMENTO}
& \(\mathbf{0.97 \pm 0.06}\)
& \(\mathbf{0.87 \pm 0.23}\)
& \(\mathbf{0.63 \pm 0.17}\)
& \(\mathbf{1.00 \pm 0.00}\)
& \(\mathbf{0.78 \pm 0.08}\) \\

SAC+HER sparse
& \(0.00 \pm 0.00\)
& --
& --
& --
& -- \\

SAC+HER dense
& \(0.00 \pm 0.00\)
& --
& --
& --
& -- \\
\bottomrule
\end{tabular}%
}
\caption{Final success rate on original and held-out generalization domains. Cube is the Robosuite Tower-of-Hanoi environment used during policy evolution. Asym. and Cyl. are held-out Robosuite object-geometry variants that preserve the Hanoi task while changing the shape of the manipulated object. Search scenes are the AI2-THOR kitchens used during policy evolution, and held-out scenes are unseen kitchens \texttt{FloorPlan4--29}. SAC+HER is evaluated only on the original Robosuite cube task. Entries are mean \(\pm\) standard deviation across three seeds when available; ``--'' denotes not evaluated.}
\label{tab:generalization}
\end{table*}

Table~\ref{tab:generalization} shows that \textsc{MEMENTO} achieves the highest final success in both training domains and maintains substantial success under domain-specific generalization tests. In Robosuite, the final policy transfers from cubic training objects to asymmetric stackable objects and cylindrical disks, indicating that the evolved program relies on Hanoi state structure, legal stack ordering, and closed-loop placement geometry rather than memorizing cube-specific contacts. In AI2-THOR, the final policy transfers from the three search kitchens to unseen kitchens with new layouts, object placements, visibility constraints, and object identifiers. REvolve partially solves the original domains, but loses most performance on the held-out Robosuite geometries and AI2-THOR kitchens, while Eureka produces no successful rollouts in either domain. These results indicate that the evolved programs rely on object-centric state assignments, legal subgoal ordering, and observation-conditioned action targets, with limited dependence on cube-specific contacts or memorized kitchen layouts.

\subsection{Ablations}
\label{sec:ablations}

The ablations use the same three-seed evaluation protocol as the main comparison. All evolutionary code-as-policy search ablations use the same total candidate budget as the main experiments, \(B=24\) evaluated policy candidates per generation. The zero-shot no-evolution baseline uses the same one-generation candidate budget of \(24\) policy candidates, without evolutionary updates. We isolate four design choices: whether evolutionary code-as-policy search is needed at all, whether an evolved evaluator is needed to guide code-as-policy search, whether the code-as-policy search branches must be used jointly, and whether restricting AI2-THOR search to a single training scene reduces cross-scene generalization.

\paragraph{Zero-shot policy generation.}
We first test whether the LLM can solve the tasks without evolutionary policy search. In this setting, we sample \(24\) zero-shot policy candidates and evaluate the best candidate, but without evolutionary generations. This baseline tests whether the domains can be solved by single-round code generation alone.

\begin{figure}[ht]
\centering
\includegraphics[width=\linewidth]{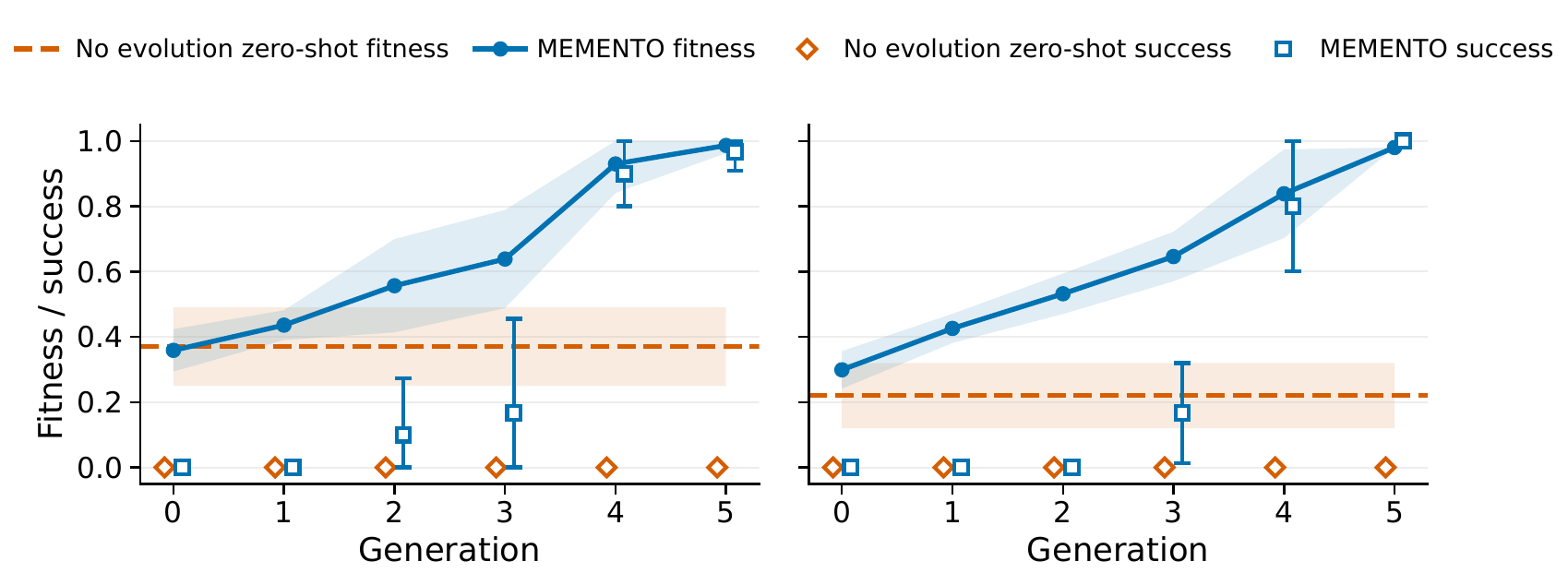}
\caption{Zero-shot policy-generation baseline compared with full \textsc{MEMENTO}. Left: Robosuite Tower-of-Hanoi. Right: AI2-THOR household interaction. The dashed horizontal curve shows the no-evolution zero-shot fitness baseline from \(24\) policy candidates, and diamond markers show its empirical success. The solid curve shows \textsc{MEMENTO} best-so-far fitness, and square markers show \textsc{MEMENTO} empirical success. Shaded regions and error bars denote one standard deviation across three seeds.}
\label{fig:zero_shot_no_evolution}
\end{figure}

On Robosuite Tower-of-Hanoi, zero-shot policy generation reaches \(0.37 \pm 0.12\) fitness and \(0.00 \pm 0.00\) success. On AI2-THOR, it reaches \(0.22 \pm 0.10\) fitness and \(0.00 \pm 0.00\) success. Thus, zero-shot policies achieve partial dense progress but never satisfy the full task-success predicate in either domain. By contrast, full \textsc{MEMENTO} reaches \(\mathbf{0.99 \pm 0.02}\) fitness and \(\mathbf{0.97 \pm 0.06}\) success on Robosuite, and \(\mathbf{0.98 \pm 0.00}\) fitness and \(\mathbf{1.00 \pm 0.00}\) success on AI2-THOR. Figure~\ref{fig:zero_shot_no_evolution} shows this comparison. The zero-shot baseline is shown as a horizontal reference because it is not an evolutionary trajectory.

\begin{figure}[ht]
\centering
\includegraphics[width=0.9\linewidth]{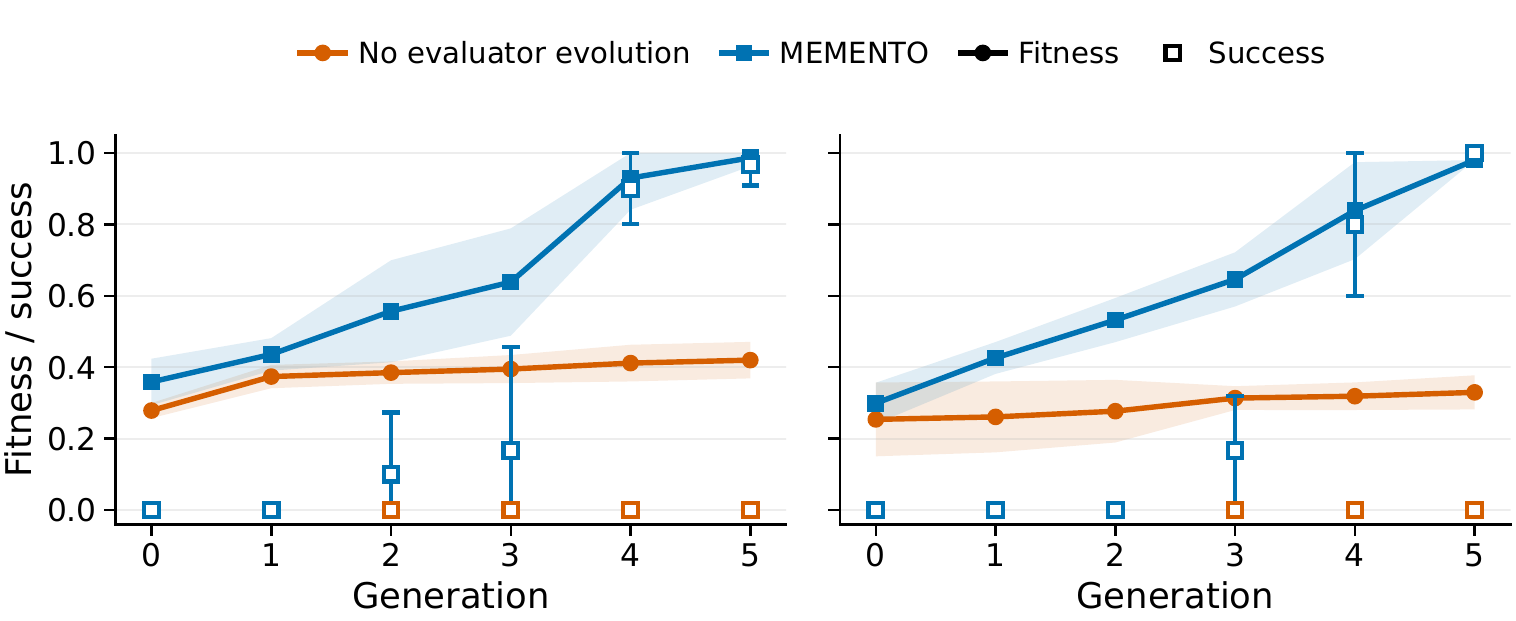}
\caption{Evaluator-evolution ablation on Robosuite Tower-of-Hanoi and AI2-THOR household interaction. Each subfigure compares full \textsc{MEMENTO} with a variant that uses one zero-shot LLM-generated evaluator per seed and does not perform evaluator evolution. Solid curves show best-so-far evaluator fitness, and square markers show empirical success across generations. Lines and markers denote means across three seeds, and shaded regions denote one standard deviation.}
\label{fig:evaluator_evolution_ablation}
\end{figure}

\paragraph{Evaluator evolution.}
We next test whether evaluator evolution is needed once policy search is run. We replace the evolved evaluator from Section~\ref{subsec:evaluator_evolution} with one zero-shot LLM-generated evaluator per seed and run the same policy-search procedure. The evaluator still returns scalar fitness and feedback metrics, but it is not mutated or selected through evaluator evolution. Figure~\ref{fig:evaluator_evolution_ablation} compares this no-evaluator-evolution variant with full \textsc{MEMENTO}.

At generation~5, the no-evaluator-evolution variant reaches
\(0.42 \pm 0.05\) fitness and \(0.00 \pm 0.00\) success on Robosuite
Tower-of-Hanoi, and \(0.33 \pm 0.05\) fitness and \(0.00 \pm 0.00\)
success on AI2-THOR. Full \textsc{MEMENTO} reaches
\(\mathbf{0.99 \pm 0.02}\) fitness and
\(\mathbf{0.97 \pm 0.06}\) success on Robosuite, and
\(\mathbf{0.98 \pm 0.00}\) fitness and
\(\mathbf{1.00 \pm 0.00}\) success on AI2-THOR. The zero-shot evaluators support limited fitness improvement but do not guide policy search to task completion. This indicates that evaluator evolution is needed to align rollout feedback with the long-horizon success conditions.

\begin{figure}[t]
\centering
\includegraphics[width=0.9\linewidth]{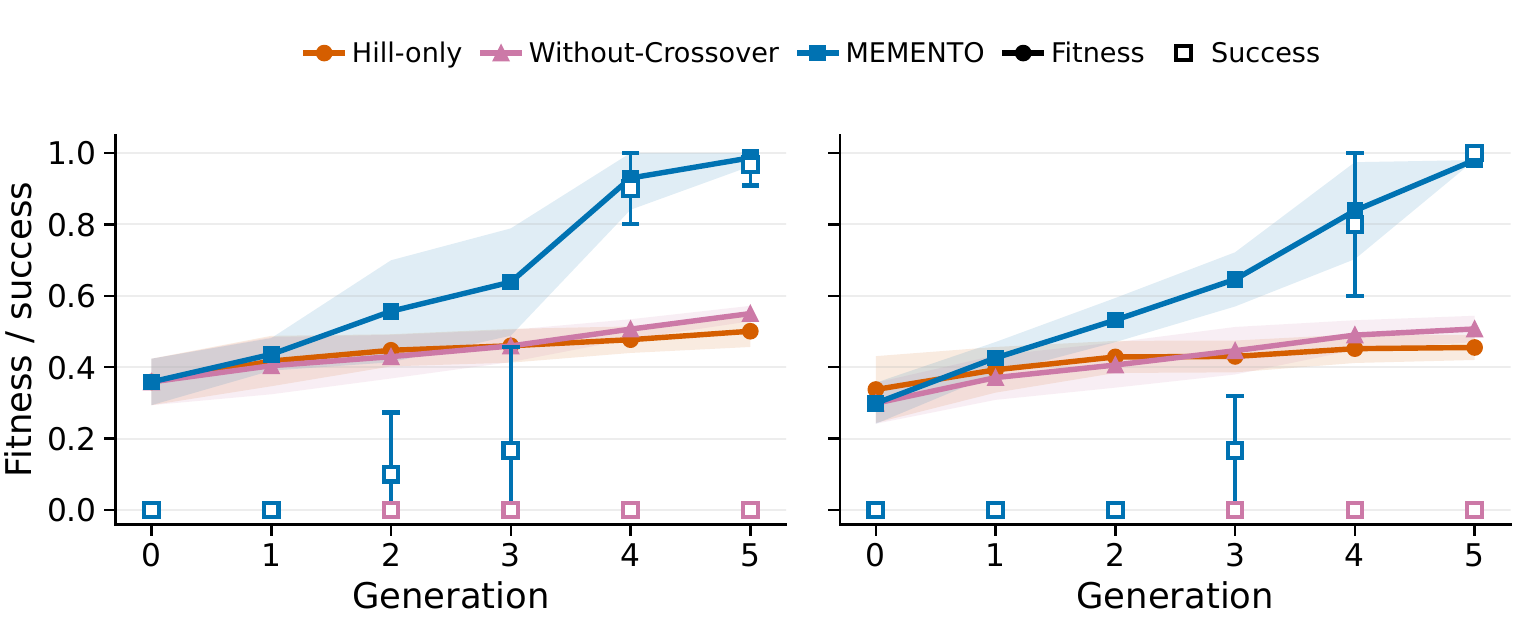}
\caption{Operator-branch ablations on Robosuite Tower-of-Hanoi and AI2-THOR household interaction. Each subfigure compares Hill-only, Without-Crossover, and full \textsc{MEMENTO}. Solid curves show best-so-far evaluator fitness, and hollow square markers show empirical success across generations. Hill-only retains sequential accepted-candidate refinement with rejected-proposal memory but removes macro-mutation and crossover. Without-Crossover retains hill-climbing and macro-mutation but removes crossover between its best candidates.}
\label{fig:operator_ablations}
\end{figure}

\paragraph{Operator branches.}
We then evaluate whether the code-as-policy search branches must be used jointly. The evolved evaluator is kept fixed, and full \textsc{MEMENTO} is compared with two ablated variants: Hill-only and Without-Crossover. Hill-only reallocates the full candidate budget to sequential hill-climbing, using \(K_{\mathrm{hc}}=24\), \(K_{\mathrm{macro}}=0\), and \(K_{\mathrm{cross}}=0\). Without-Crossover removes recombination and reallocates the crossover budget to macro-mutation, using \(K_{\mathrm{hc}}=12\), \(K_{\mathrm{macro}}=12\), and \(K_{\mathrm{cross}}=0\). A macro-mutation-only variant is not included because it matches the mutation-only structure of the adapted Eureka baseline. Figure~\ref{fig:operator_ablations} reports the resulting fitness and success curves.

At generation~5, Hill-only reaches \(0.50 \pm 0.04\) fitness on Robosuite Tower-of-Hanoi and \(0.46 \pm 0.04\) on AI2-THOR, while Without-Crossover reaches \(0.55 \pm 0.02\) and \(0.51 \pm 0.04\), respectively. Neither ablated variant produces successful rollouts in either domain. In contrast, full \textsc{MEMENTO} reaches
\(\mathbf{0.99 \pm 0.02}\) fitness and
\(\mathbf{0.97 \pm 0.06}\) success on Robosuite, and
\(\mathbf{0.98 \pm 0.00}\) fitness and
\(\mathbf{1.00 \pm 0.00}\) success on AI2-THOR. These results indicate that, under the tested candidate budget, local refinement and macro-mutation alone are insufficient: the crossover branch is needed for \textsc{MEMENTO} to combine partial executable strategies into task-completing policies.

\paragraph{Search-scene diversity.}
Finally, we test whether AI2-THOR policy evolution requires multiple training kitchens. In this ablation, \textsc{MEMENTO} is restricted to a single training kitchen during policy evolution and is then evaluated across \texttt{FloorPlan1}--\texttt{FloorPlan29}. The resulting policies solve the search scene but generalize poorly across kitchens, reaching \(0.09 \pm 0.02\) success across seeds \((2/29, 3/29, 3/29)\). This suggests that single-scene policy evolution overfits to scene-specific object placements, navigation routes, and interaction viewpoints. In contrast, multi-scene policy evolution exposes the policy search to scene-dependent object identifiers, visibility constraints, receptacle locations, and valid-action preconditions. Together with the held-out scene results in Table~\ref{tab:generalization}, this shows that search-scene diversity is necessary for AI2-THOR cross-scene generalization.

\subsection{Sim-to-Real Transfer}
\label{sec:sim2real}

The physical deployment tests whether the evolved program logic completes the task when its object-centric inputs are estimated from real RGB-D perception, and its selected moves are executed through physical contact. The best-performing Robosuite Tower-of-Hanoi code-as-policy across the three \textsc{MEMENTO} seeds, denoted \(p_{\mathrm{best}}\), was deployed on a physical Franka robot. The evolved task-level code is kept fixed. During the simulation search, the selected policy acquired a closed-loop execution structure that reduces dependence on exact simulator dynamics: object poses are filtered before target computation, pick-and-place targets are re-derived from the current cube and container estimates, and contact-sensitive execution is guarded by alignment checks, placement centering, grasp verification, and post-placement retry. The symbolic Hanoi ordering remains fixed, while the execution of each scheduled move is adapted to the currently estimated configuration of objects. Additional details of the evolved Hanoi policy are given in Appendix~\ref{app:best_evolved_franka_policy}.

Simulator object states, and end-effector control increments are replaced by RGB-D object-pose estimates and Franka motion commands. During the simulation search, observation noise and calibration bias were sampled from the randomization model in Appendix~\ref{app:franka_heavy_dr}. In the physical trials, analogous errors arise from segmentation, depth measurement, camera calibration, and contact execution. In simulation, Robosuite provides object poses and robot state, and actions are applied through the Robosuite end-effector controller. In the physical setup, object and container poses are estimated from RGB-D observations using SAM3 segmentation \citep{carion2025sam}, and policy-selected pick-and-place targets are executed by the robot motion planner and gripper controller. 

Before the final robot evaluation, a small set of execution constants was recalibrated for the physical cell over \(N_{\mathrm{cal}}=3\) LLM-assisted calibration steps. The calibration variables, real-robot metrics, and calibration procedure are detailed in Appendix~\ref{app:real_world_deployment}. The real-robot metrics correspond to a subset of the evaluator's feedback categories and are computed by a reduced evaluator \(E_R\). They guide calibration only; no fitness is computed, and no policy search is performed during deployment.

The calibrated code-as-policy achieved a success rate of \(0.90\) over \(N_{\mathrm{real}}=10\) physical trials. In all \(10\) trials, the task-level code followed the fixed Hanoi ordering and rederived execution targets from the estimated object configuration; the single unsuccessful trial was due to a perception error.

\section{Limitations and Future Work}

The experiments use two task formulations: four-cube Tower-of-Hanoi manipulation in Robosuite and a long-horizon household-interaction task in AI2-THOR. Since each evolved program is conditioned on a task-specific prompt, evaluator, and success predicate, broader task coverage would require running the full evaluator-evolution and code-as-policy search procedure on additional task definitions. Model sensitivity is not isolated, as evaluator and policy proposals are generated using a single language model. The evaluator \(E^\star\) is evolved from the initial-policy rollout set \(\mathcal{T}_0\) and then fixed during code-as-policy search, so the experiments do not test whether evaluator evolution should be reintroduced when later policy elites expose failures absent in the initial rollouts. The hill-climb branch is sequential within each generation because each proposal depends on the currently accepted candidate and the rejection memory, thereby limiting within-branch parallelism relative to population-based variation.

Future work should therefore extend the current single-elite formulation to a population-based code-as-policy search. Multiple islands could maintain separate policy elites, run branch-level search independently, and periodically exchange high-fitness candidates through migration~\citep{island_ea_seminal}. This would add population diversity and reduce the sequential bottleneck of a single hill-climb chain. In addition, future work should also study adaptive evaluator--policy co-evolution, in which \(E^\star\) is periodically re-evolved from rollouts of later elites or from failed and ambiguous candidates collected during the search, rather than only from \(p_0\). Finally, evaluator and policy generation with multiple language models should also be tested to measure model sensitivity.

\section{Conclusion}
\label{sec:conclusion}

This work introduced \textsc{MEMENTO}, a memory-guided single-elite memetic framework for evolving executable code-as-policy programs in long-horizon embodied tasks. The method first evolves a rollout evaluator that maps executions to scalar fitness and task-specific feedback metrics, then uses the fixed evaluator to guide policy search through sequential hill-climbing with rejected-proposal memory, macro-mutation, and crossover between the best hill-climb and macro-mutation candidates. On the Robosuite Franka Tower-of-Hanoi and AI2-THOR household interaction tasks, \textsc{MEMENTO} achieved higher performance than LLM-guided evolutionary baselines adapted to code-as-policy search, including on held-out object geometries and unseen household scenes. Ablations showed that zero-shot policy generation, policy search with an unevolved zero-shot evaluator, and removal of search branches each reduced final performance. Finally, the best evolved Robosuite Tower-of-Hanoi code-as-policy was deployed on a physical Franka robot. These results support memory-guided memetic search as a mechanism to improve executable embodied policies using sparse rollout feedback.

\small

\bibliographystyle{apalike}
\bibliography{references}

\clearpage
\appendix

The appendix is organized as follows.
Appendix~\ref{app:franka_env} details the Robosuite Franka
Tower-of-Hanoi environment, domain randomization, the physical Franka
deployment, held-out object variants, and the SAC+HER baselines.
Appendix~\ref{app:ai2thor_env} details the AI2-THOR environment, the
code-as-policy interface, and the observation space.
Appendix~\ref{app:additional_results} reports the evolved evaluators
and representative evolved policy programs.

\section{Robosuite Franka Tower-of-Hanoi Environment}
\label{app:franka_env}

\subsection{Task and Simulator Setup}
\label{app:franka_task}

The Robosuite task uses a Franka Panda arm, four cubes, and three containers. The cubes are ordered by size: extra-large, large, medium, and small. At reset, all four cubes are stacked on container A in legal Tower-of-Hanoi order, with the extra-large cube at the bottom and the small cube at the top. The goal is to move the complete tower to container C, preserve the same size order, and leave containers A and B empty. An episode terminates on success, illegal Hanoi placement, or timeout after 12000 environment steps.

\begin{figure}[ht]
\centering
\includegraphics[
width=1\linewidth,
trim={0 0 0 30pt},
clip
]{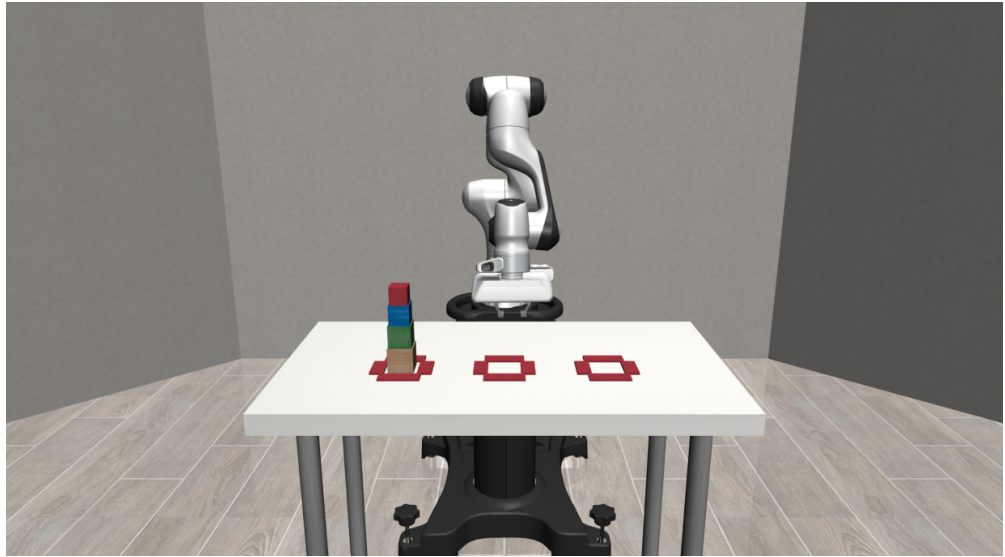}
\caption{Initial reset state of the Robosuite Franka Panda Tower-of-Hanoi task. Four cubes are stacked on container A; the target is to move the full tower to container C while preserving the size order.}
\label{fig:franka_hanoi_initial_state}
\end{figure}
The cube half-edge lengths are fixed by the environment:
\[
s_{\mathrm{small}}=0.022,\quad
s_{\mathrm{medium}}=0.025,\quad
s_{\mathrm{large}}=0.028,\quad
s_{\mathrm{xlarge}}=0.031.
\]
The containers are plate-with-hole objects. The hole is shifted from the container origin along the container-local positive \(x\)-axis. The local hole interval is \[
x_{\mathrm{local}}\in[0.02,0.09],
\qquad
|y_{\mathrm{local}}|\leq 0.035,
\] so the nominal hole center is at local offset
\[
h_{\mathrm{local}}=[0.055,0,0]^\top.
\]
For a container with world position \(b\) and orientation \(R_b\), the placement center used by policy code is
\[
h(b,R_b)=b+R_b h_{\mathrm{local}}.
\]

\subsection{Observation and Action Space}
\label{app:franka_cap_obs_action}

Policy-code evolution generates a Python class \texttt{FrankaPolicy} with two required methods. The method \texttt{reset()} clears episode-level policy state before each rollout. The method \texttt{compute\_action(obs)} is called at every control step and returns a continuous seven-dimensional action vector:
\[
a_t =
[\Delta x,\Delta y,\Delta z,\Delta\phi,\Delta\theta,\Delta\psi,\Delta g]
\in[-1,1]^7.
\]
The first three entries are end-effector position increments, the next three entries are axis-angle orientation increments, and the final entry controls the gripper. Positive gripper values close the gripper, negative values open it, and zero holds the current command.

The generated policy receives the structured Robosuite observation dictionary directly. The task-relevant observation fields are listed in Table~\ref{tab:franka_cap_obs}. The policy is instructed to use these structured keys rather than packed aggregate vectors.

\begin{table}[h]
\centering
\small
\caption{Observation fields used by policy-code evolution in the Robosuite Franka Tower-of-Hanoi task.}
\label{tab:franka_cap_obs}
\begin{tabular}{lll}
\toprule
Group & Key & Content \\
\midrule
Robot & \texttt{robot0\_eef\_pos} & End-effector position, shape \((3,)\) \\
Robot & \texttt{robot0\_eef\_quat} & End-effector quaternion, shape \((4,)\) \\
Robot & \texttt{robot0\_gripper\_qpos} & Gripper joint positions, shape \((2,)\) \\
Robot & \texttt{robot0\_gripper\_qvel} & Gripper joint velocities, shape \((2,)\) \\
Robot & \texttt{robot0\_joint\_pos} & Arm joint positions, shape \((7,)\) \\
Robot & \texttt{robot0\_joint\_vel} & Arm joint velocities, shape \((7,)\) \\
\midrule
Cubes & \texttt{cube\_pos} & Four cube positions, shape \((12,)\) \\
Cubes & \texttt{cube\_quat} & Four cube quaternions, shape \((16,)\) \\
Cubes & \texttt{cube\_size} & Four cube half-edge lengths, shape \((4,)\) \\
\midrule
Containers & \texttt{boxes\_pos} & Container A/B/C positions, shape \((9,)\) \\
Containers & \texttt{boxes\_quat} & Container A/B/C quaternions, shape \((12,)\) \\
\bottomrule
\end{tabular}
\end{table}

The evolved policy parses the four cube poses and sizes, computes the hole center of each container from \texttt{boxes\_pos} and \texttt{boxes\_quat}, and infers the current cube-to-container assignment. Its symbolic component generates the recursive Tower-of-Hanoi move sequence from container \(A\) to container \(C\) through container \(B\), yielding \(2^4-1=15\) ordered moves. For each scheduled move, the policy selects the top observed cube on the source container, recomputes pick-and-place targets from the current object and container poses, and executes the move via closed-loop pick-and-place control. The symbolic Hanoi ordering remains fixed; closed-loop adaptation comes from observation-based cube assignment, grasp/lift checks, and geometric target recomputation during execution.

\subsection{Domain Randomization for Policy-Code Evolution}
\label{app:franka_heavy_dr}

At each reset, the environment randomizes observation noise, calibration bias, contact parameters, cube mass, joint damping, actuator gain, world position offset, and initial cube placement. These perturbations modify both the object-state observations received by the policy and the underlying MuJoCo dynamics, while preserving the same task definition and success condition.

\begin{table}[t]
\centering
\small
\caption{Heavy domain randomization used for policy-code evolution in the Robosuite Franka Tower-of-Hanoi environment.}
\label{tab:franka_heavy_dr}
\begin{tabular}{ll}
\toprule
Quantity & Value \\
\midrule
Sliding friction scale & \([0.3,1.8]\) \\
Torsional friction & \([0.001,0.01]\) \\
Rolling friction & \([0.0001,0.001]\) \\
Position-noise standard deviation & \([0.002,0.025]\) \\
Position bias & \(\pm 0.015\) \\
World XY offset & \(\pm 0.015\) \\
Cube mass scale & \([0.7,1.5]\) \\
Joint damping scale & \([0.8,1.5]\) \\
Actuator gain scale & \([0.85,1.15]\) \\
Initial cube XY perturbation & \(\pm 0.003\) \\
\bottomrule
\end{tabular}
\end{table}

The position-noise standard deviation is sampled once per episode and then used when cube and container positions are observed. The resulting sensor value is the true object position plus an episode-level position bias and per-observation Gaussian noise. The world XY offset changes the physical placement of containers A, B, and C at reset. The initial cube XY perturbation applies an independent physical offset to each cube in the initial tower. Observation latency is not included in Table~\ref{tab:franka_heavy_dr}, because it is not applied to the returned observations in the implementation used for the experiments.

\subsection{Physical Franka Deployment}
\label{app:real_world_deployment}

This appendix details the physical Franka deployment used for the sim-to-real result in Section~\ref{sec:sim2real}. The deployed policy is the best-performing Robosuite Tower-of-Hanoi code-as-policy selected across the MEMENTO seeds. The selected policy uses a fixed, recursive Tower-of-Hanoi move schedule, together with guarded, closed-loop execution of each pick-and-place move. The evolved program logic is kept fixed, while the simulator observation and action interfaces are replaced by real-robot perception and execution.

\paragraph{Deployment interface.}
Let \(p_{\mathrm{best}}\) denote the selected Robosuite code-as-policy. In simulation, \(p_{\mathrm{best}}\) receives the Robosuite observation dictionary \(o_t\), including robot state, cube poses, cube sizes, and container poses, and returns a seven-dimensional end-effector action. On the physical robot, a deployment wrapper converts RGB-D observations and robot-state measurements into the same object-centric fields consumed by the policy:
\[
o_t^R = \Psi_R(y_t^R),
\]
where \(y_t^R\) denotes the raw real-robot observation and \(\Psi_R\) estimates the cube poses, cube sizes, container poses, and end-effector pose in the Franka base frame.

Given \(o_t^R\), the physical execution version of the policy evaluates the same internal finite-state controller as in simulation. At each control step, the policy identifies the active manipulation phase, computes the current Cartesian pick or place target, and issues the corresponding gripper command:
\[
(\alpha_t,x_t^\star,g_t,z_{t+1})
=
p_{\mathrm{best}}^R(o_t^R,z_t).
\]
Here \(\alpha_t\) is the active manipulation phase, \(x_t^\star\) is the Cartesian target, \(g_t\) is the gripper command, and \(z_t\) is the policy state. The physical execution stack maps these phase-specific targets to Franka motion and gripper commands,
\[
u_t^R = \Gamma_R(\alpha_t,x_t^\star,g_t).
\]
Thus, \(\Psi_R\) replaces the simulator observation source and \(\Gamma_R\) replaces the simulator action interface, while the Tower-of-Hanoi state machine remains fixed.

\paragraph{Preserved program logic.}
The preserved policy components are the object-centric parts of the evolved code-as-policy. These include the recursive \(15\)-move Tower-of-Hanoi schedule, the active move index, the manipulation finite-state machine, filtered pose estimates, retry-conditioned pick offsets, guarded descent, stack-height-based placement-target computation, grasp verification, placement centering, and post-placement verification. The deployed policy does not perform global BFS replanning. Instead, the symbolic Hanoi ordering remains fixed, and recovery is local to the active move: failed grasps or failed post-placement checks cause the same scheduled move to be retried from the updated object observation.

\paragraph{RGB-D object grounding.}
Robosuite object-state observations are replaced by RGB-D object grounding. SAM3-based language-guided segmentation is queried with the physical object labels used in the task: the four cube colors and the three container markers \citep{carion2025sam}. The resulting segmentation masks are combined with depth measurements and calibrated camera extrinsics to estimate object poses in the Franka base frame. The estimated cube poses, cube sizes, container poses, and end-effector pose are written into the same object-centric fields consumed by \(p_{\mathrm{best}}\). Cube poses are estimated online during execution. Container poses are measured once and reused as fixed container locations during the physical trials.

\paragraph{Physical execution.}
Policy-selected pick-and-place targets are executed by the Franka motion and gripper stack. Long transfers between task locations are executed through motion planning. Short vertical motions for grasping, lifting, placing, and retreating are executed as local Cartesian motions. The gripper command is provided by the policy and executed by the physical wrapper. After each executed phase, the current object configuration is observed again, and the active move targets are recomputed from the updated object-centric state. If the grasp verification or post-placement checking fails, the policy retries the same scheduled Hanoi move rather than advancing the symbolic move index.

\paragraph{Reduced real-robot evaluator.}
The simulator evaluator \(E^\star\) returns scalar fitness and feedback metrics from simulator rollouts,
\[
E^\star(\mathcal{T}_{N_{\mathrm{roll}}}(p))
=
\bigl(F(p),\mathbf{M}(p)\bigr).
\]
For physical deployment, a reduced evaluator \(E_R\), distinct from the evolved \(E^\star\), computes feedback metrics only from robot execution and returns no fitness. Let \(\tau_i^R(p_{\mathrm{best}},\kappa_i)\) denote the physical execution obtained at calibration step \(i\) with execution constants \(\kappa_i\). The reduced evaluator is
\[
E_R
\bigl(\tau_i^R(p_{\mathrm{best}},\kappa_i)\bigr)
=
\mathbf{M}_i^R,
\qquad
\mathbf{M}_i^R =
\bigl(m_{i,1}^R,\ldots,m_{i,d_R}^R\bigr).
\]
The metrics in \(\mathbf{M}_i^R\) are measured from robot execution and perceived object poses. They are used to calibrate the deployed controller. \(E_R\) does not introduce a new policy-search objective and is not used for selection during MEMENTO search.

\paragraph{LLM-assisted calibration.}
Calibration adjusts execution constants before the final physical evaluation. Let \(\kappa_i\) denote the execution constants at calibration step \(i\), and let \(c_R\) denote the real-robot execution code. At each step, the current deployment is executed on the robot, \(\mathbf{M}_i^R\) is computed by \(E_R\), and the language model receives
\[
\left(
p_{\mathrm{best}},\;
c_R,\;
\kappa_i,\;
\mathbf{M}_i^R
\right).
\]
The calibration proposal is
\[
\kappa_{i+1}
=
\mathcal{L}_{\mathrm{cal}}
\left(
p_{\mathrm{best}},
c_R,
\kappa_i,
\mathbf{M}_i^R
\right),
\qquad
i=0,\ldots,N_{\mathrm{cal}}-1 .
\]
We use \(N_{\mathrm{cal}}=3\) calibration steps. The final constants \(\kappa^\star\) are selected from the tested calibration sequence and used for the final physical evaluation.

\paragraph{Calibrated constants.}
Table~\ref{tab:real_world_calibration_constants} reports the execution constants changed for physical deployment. These constants affect pick depth, approach height, container assignment tolerance, and place-hover height. 

\begin{table}[h]
\centering
\small
\caption{Execution constants for physical Franka deployment. Values are in meters.}
\label{tab:real_world_calibration_constants}
\begin{tabular}{lcc}
\toprule
Parameter & Simulation & Physical robot \\
\midrule
\texttt{grasp\_depth\_offset}      & \(0.002\) & \(0.045\) \\
\texttt{pick\_pregrasp\_clearance} & \(0.040\) & \(0.090\) \\
\texttt{peg\_assign\_xy}           & \(0.055\) & \(0.075\) \\
\texttt{place\_hover\_clearance}   & \(0.110\) & \(0.100\) \\
\bottomrule
\end{tabular}
\end{table}

\paragraph{Physical target corrections.}
The physical execution wrapper also applies small Cartesian corrections after the policy selects the current pick-or-place target. Pick targets are shifted by a fixed lateral centering offset. Stack-placement targets receive a fixed lateral correction when placed onto an existing stack. During the place-descend phase, the target height is lowered to seat the cube on the support surface. These corrections are part of the real execution interface and do not alter move selection, cube-to-container assignment, or move retry.

\subsection{Held-Out Tower-of-Hanoi Object Variants}
\label{app:gen_hanoi}

To evaluate object-geometry transfer, we use two held-out Robosuite Tower-of-Hanoi variants. Both variants preserve the Hanoi objective, object identifiers, target layout, and structured policy observation interface while changing the geometry of the manipulated objects. This tests whether the policy relies on Hanoi state relations, such as target assignment, stack order, and placement targets, rather than on the cubic object geometry used during policy search.

\begin{figure}[t]
\centering
\begin{minipage}{0.48\linewidth}
\centering
\includegraphics[width=\linewidth]{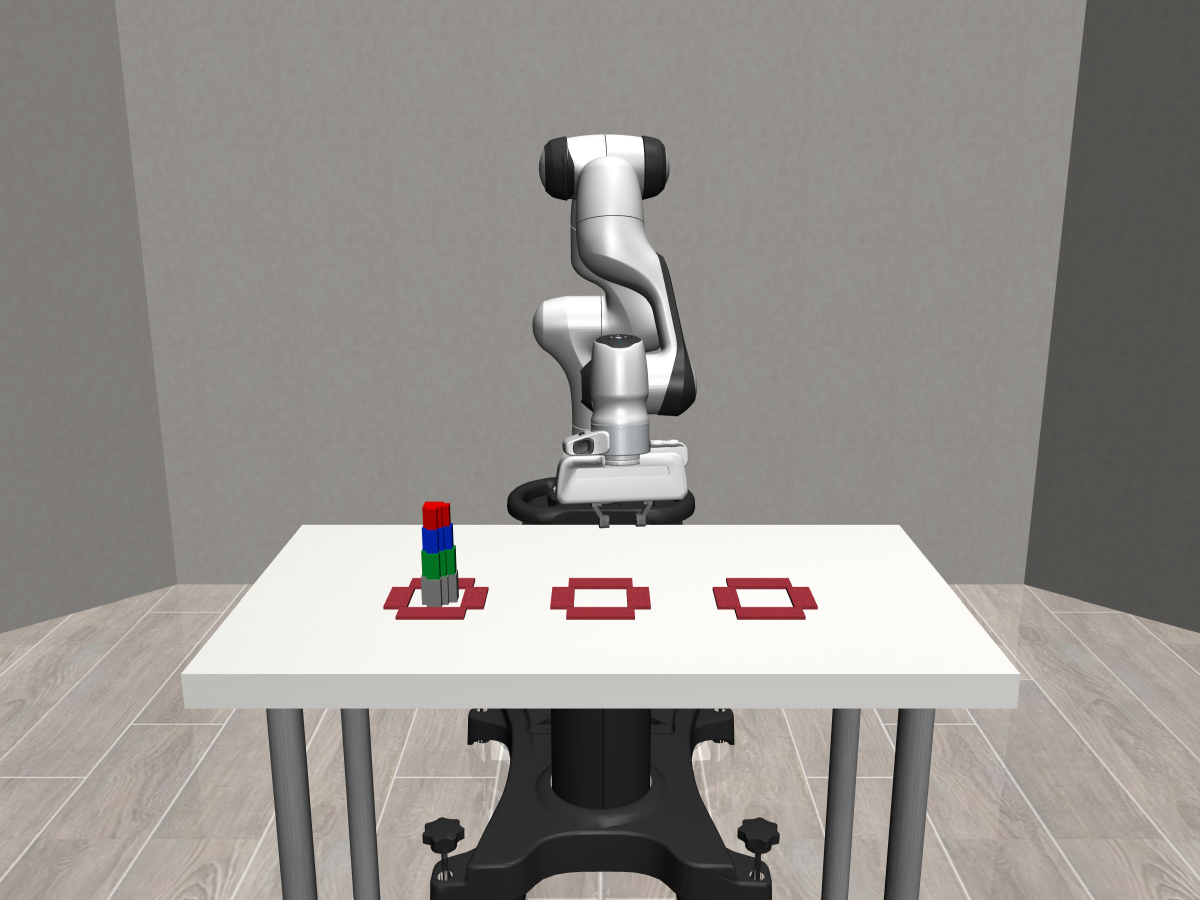}
\vspace{-0.4em}
\centerline{\small (a) Asymmetric stackable objects}
\end{minipage}
\hfill
\begin{minipage}{0.48\linewidth}
\centering
\includegraphics[width=\linewidth]{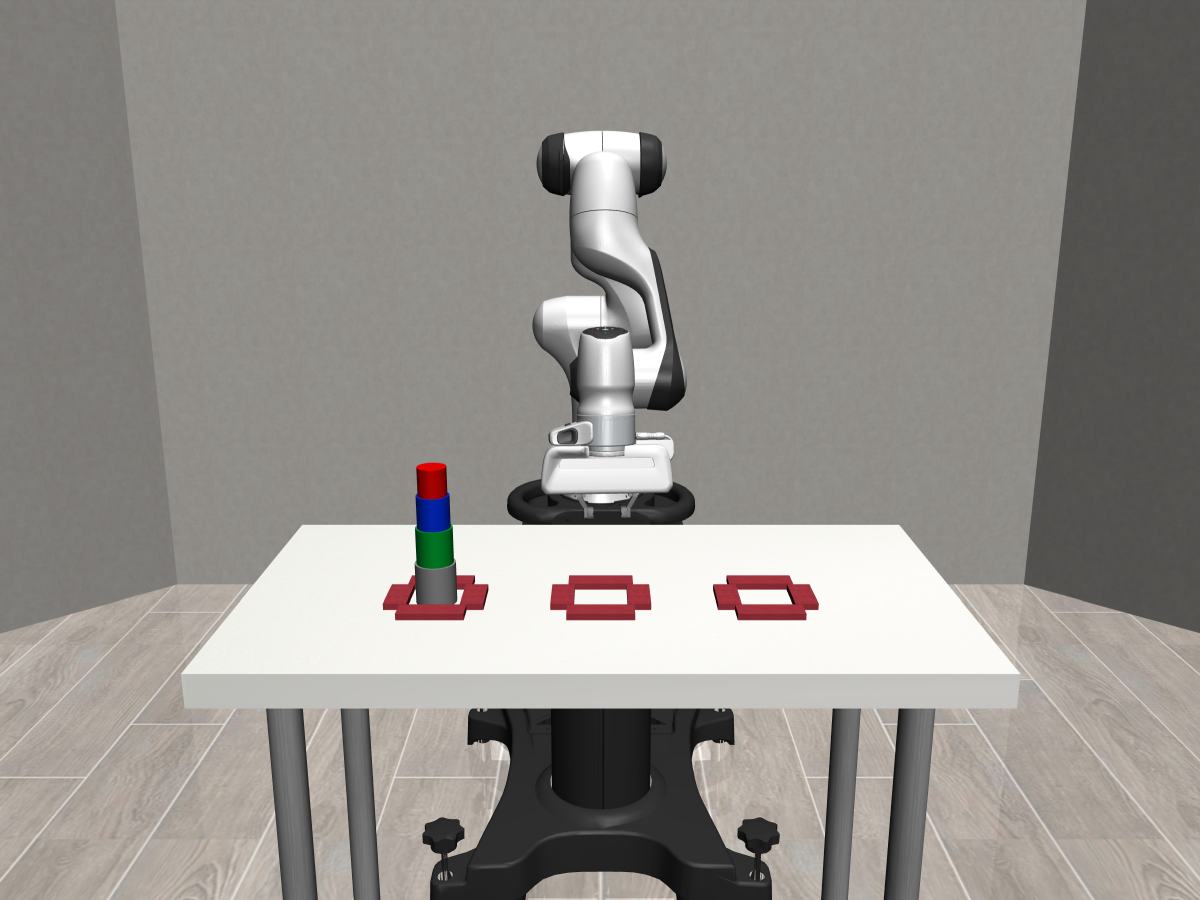}
\vspace{-0.4em}
\centerline{\small (b) Cylindrical disks}
\end{minipage}
\caption{Held-out Robosuite Tower-of-Hanoi object variants. Both variants preserve the Hanoi task and policy observation interface used for the cubic training environment. The asymmetric-object variant changes the grasp envelope while retaining a stackable support core. The cylindrical-disk variant changes the object contact geometry by replacing cube faces with curved cylinder sides.}
\label{fig:gen_hanoi_variants}
\end{figure}

The asymmetric-object variant replaces each cube with a stackable central core and two protruding arms. The object remains compatible with the Hanoi stacking task, but its side surfaces and grasp points differ from those of a cube with the same scalar size. The cylindrical-disk variant replaces each cube with a cylindrical disk. This keeps flat top--bottom stacking surfaces but changes the lateral contact geometry during grasping and transport.

\begin{table}[h]
\centering
\small
\caption{Difference between the training Robosuite Tower-of-Hanoi
environment and the held-out object variants.}
\label{tab:gen_hanoi_variants}
\begin{tabular}{lll}
\toprule
Environment & Object geometry & Main change tested \\
\midrule
Cube Hanoi &
Symmetric cubes &
Training object family \\
Asymmetric Hanoi &
Stackable core with protruding arms &
Non-cubic grasp envelope \\
Cylinder Hanoi &
Cylindrical disks &
Curved side contacts \\
\bottomrule
\end{tabular}
\end{table}

\subsection{SAC+HER Observation and Action Space}
\label{app:franka_sac_obs_action}

The SAC+HER baselines use the same Robosuite task and continuous low-level action space as the policy-code experiments, but their observation representation differs. The HER wrapper returns a Gymnasium dictionary observation:
\[
o_t =
\left\{
o_t^{\mathrm{base}},
g_t^{\mathrm{ach}},
g^\star
\right\}.
\]
The base observation \(o_t^{\mathrm{base}}\) is the flattened Robosuite object-state observation. The achieved goal \(g_t^{\mathrm{ach}}\) is the current Cartesian position of the four cubes,
\[
g_t^{\mathrm{ach}}
=
[
x_t^{\mathrm{small}},
x_t^{\mathrm{medium}},
x_t^{\mathrm{large}},
x_t^{\mathrm{xlarge}}
]
\in\mathbb{R}^{12}.
\]
The desired goal \(g^\star\) is the target tower on container C, with the extra-large cube at the bottom, followed by the large, medium, and small cubes. Quaternion goals are disabled, so both achieved and desired goals contain cube positions only.

The use of \texttt{MultiInputPolicy} follows from this dictionary observation. SAC receives \texttt{observation}, \texttt{achieved\_goal}, and \texttt{desired\_goal} as separate input fields. The action space is the Robosuite continuous action box exposed by the Franka environment.

A goal is successful when all four cube positions are within
\(\epsilon_{\mathrm{goal}}=0.025\) of their corresponding desired
positions:
\[
\mathrm{succ}(g_t^{\mathrm{ach}},g^\star)
=
\mathbb{I}
\left[
\max_{j\in\{1,\ldots,4\}}
\left\|
g_{t,j}^{\mathrm{ach}}-g_j^\star
\right\|_2
<
\epsilon_{\mathrm{goal}}
\right].
\]

\subsection{SAC+HER Training Configuration}
\label{app:franka_sac_training}

Both SAC+HER baselines use Soft Actor-Critic with Hindsight Experience Replay. HER uses future-goal relabeling: additional goals are sampled from states achieved later in the same trajectory, and the reward is recomputed using the sparse or dense goal-conditioned reward.

\begin{table}[t]
\centering
\small
\caption{Shared SAC+HER training configuration for the Robosuite Franka
Tower-of-Hanoi task.}
\label{tab:sac_her_config}
\begin{tabular}{lll}
\toprule
Group & Parameter & Value \\
\midrule
Environment & Robot & Panda \\
Environment & Camera observations & disabled \\
Environment & Object observations & enabled \\
Environment & Control frequency & \(20\) Hz \\
Environment & Episode horizon & \(12000\) steps \\
Environment & Goal tolerance & \(0.025\) \\
\midrule
SAC & Policy & MultiInputPolicy \\
SAC & Total timesteps & \(50{,}000{,}000\) \\
SAC & Learning rate & \(10^{-4}\) \\
SAC & Replay buffer size & \(2{,}000{,}000\) \\
SAC & Learning starts & \(25{,}000\) \\
SAC & Batch size & \(512\) \\
SAC & Discount factor \(\gamma\) & \(0.995\) \\
SAC & Target smoothing \(\tau\) & \(0.005\) \\
SAC & Train frequency & \(1\) step \\
SAC & Gradient steps & \(1\) \\
SAC & Entropy coefficient & auto \\
\midrule
HER & Replay buffer & HerReplayBuffer \\
HER & Goal selection strategy & future \\
HER & Sampled goals per transition & \(8\) \\
\midrule
Evaluation & Evaluation frequency & \(1{,}000{,}000\) steps \\
Evaluation & Evaluation episodes & \(5\) \\
\bottomrule
\end{tabular}
\end{table}

Episode success, episode fitness, phase scores, milestones, and interaction-quality metrics are logged at episode termination. These metrics are used for analysis and comparison with policy-code evolution. The SAC update uses only the HER reward defined in Appendix~\ref{app:franka_sparse_sac_her} and Appendix~\ref{app:franka_dense_sac_her}. The SAC+HER baselines use the goal-conditioned wrapper described above; they do not use the heavy domain-randomized environment from Appendix~\ref{app:franka_heavy_dr}.

\subsection{Sparse SAC+HER}
\label{app:franka_sparse_sac_her}

The sparse SAC+HER baseline uses a success-only goal-conditioned reward:
\[
r_{\mathrm{sparse}}(g_t^{\mathrm{ach}},g^\star)
=
\begin{cases}
0, & \mathrm{succ}(g_t^{\mathrm{ach}},g^\star)=1,\\
-1, & \mathrm{succ}(g_t^{\mathrm{ach}},g^\star)=0.
\end{cases}
\]
All unsuccessful configurations receive the same reward, independent of their distance to the target tower or their intermediate stack structure. During HER replay, relabeled transitions are scored by the same success-only criterion with respect to the sampled future goal.

\subsection{Dense SAC+HER}
\label{app:franka_dense_sac_her}

The dense SAC+HER baseline uses the same observation representation, action space, SAC configuration, and HER relabeling method, but replaces the success-only reward with geometric shaping. Let
\[
d_{t,j}
=
\left\|
g_{t,j}^{\mathrm{ach}}-g_j^\star
\right\|_2
\]
be the Cartesian position error for cube \(j\), and let
\[
\bar d_t
=
\frac{1}{4}
\sum_{j=1}^{4} d_{t,j}
\]
be the mean cube-position error. The dense reward is
\[
r_{\mathrm{dense}}(g_t^{\mathrm{ach}},g^\star)
=
-\bar d_t
+
5\,\mathrm{succ}(g_t^{\mathrm{ach}},g^\star)
-
\left(1-\mathrm{legal}(g_t^{\mathrm{ach}})\right).
\]
The legality term checks the vertical cube ordering:
\[
\mathrm{legal}(g_t^{\mathrm{ach}})
=
\mathbb{I}
\left[
z_t^{\mathrm{xlarge}}
<
z_t^{\mathrm{large}}
<
z_t^{\mathrm{medium}}
<
z_t^{\mathrm{small}}
\right].
\]
The dense reward combines distance-to-target shaping, an exact success bonus, and a penalty for violating the Tower-of-Hanoi ordering constraint. During HER replay, the same dense reward is recomputed after the desired goal is replaced by the sampled future achieved goal.

\subsection{Sparse--Dense Difference}
\label{app:franka_sparse_dense_difference}

The sparse and dense SAC+HER baselines differ only in the goal-conditioned reward returned by \texttt{compute\_reward}. The sparse baseline uses only final goal achievement. The dense baseline adds continuous geometric progress and an ordering penalty. Both baselines use the same goal-conditioned observation representation, continuous action space, HER relabeling method, SAC architecture, training schedule, and logging metrics.

\section{AI2-THOR Household Environment}
\label{app:ai2thor_env}

\subsection{Task and Simulator Setup}
\label{app:ai2thor_task}

The AI2-THOR domain is a partially observed household interaction task in kitchen scenes. The task is \texttt{apple\_in\_microwave\_on\_bread\_in\_fridge}. The agent must find an Apple, a Bread, a Microwave, and a Fridge; place the Apple in the
Microwave, toggle the Microwave on, and place the Bread in the Fridge.
Success is defined by
\[
\mathrm{apple\_in\_microwave}
\;\wedge\;
\mathrm{microwave\_on}
\;\wedge\;
\mathrm{bread\_in\_fridge}.
\]
The task is partially observed: the agent receives egocentric observations, and only currently visible objects are exposed for interaction. Object identifiers are scene-specific and must be obtained from the observation before they can be used in object actions.

\begin{figure}[t]
\centering
\begin{tabular}{ccccc}
\begin{minipage}{0.28\linewidth}
  \centering
  \includegraphics[width=\linewidth,trim=0 0 0 20pt,clip]{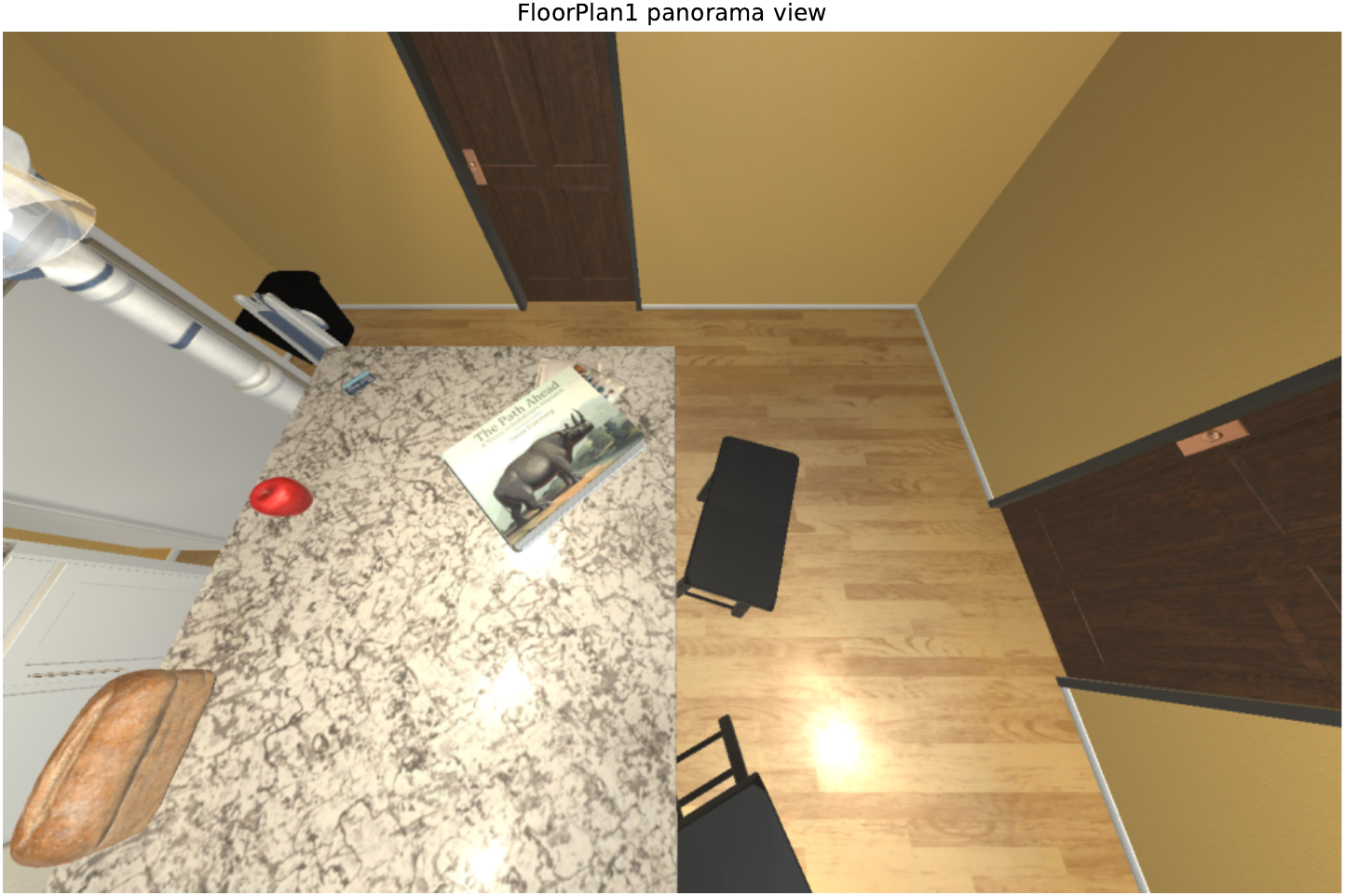}\\[-0.2em]
  {\small \texttt{FloorPlan1}}
\end{minipage}
&
\begin{minipage}{0.28\linewidth}
  \centering
  \includegraphics[width=\linewidth,trim=0 0 0 20pt,clip]{images/FloorPlan3_agent_front.pdf}\\[-0.2em]
  {\small \texttt{FloorPlan3}}
\end{minipage}
&
{\Large \(\cdots\)}
&
\begin{minipage}{0.28\linewidth}
  \centering
  \includegraphics[width=\linewidth,trim=0 0 0 20pt,clip]{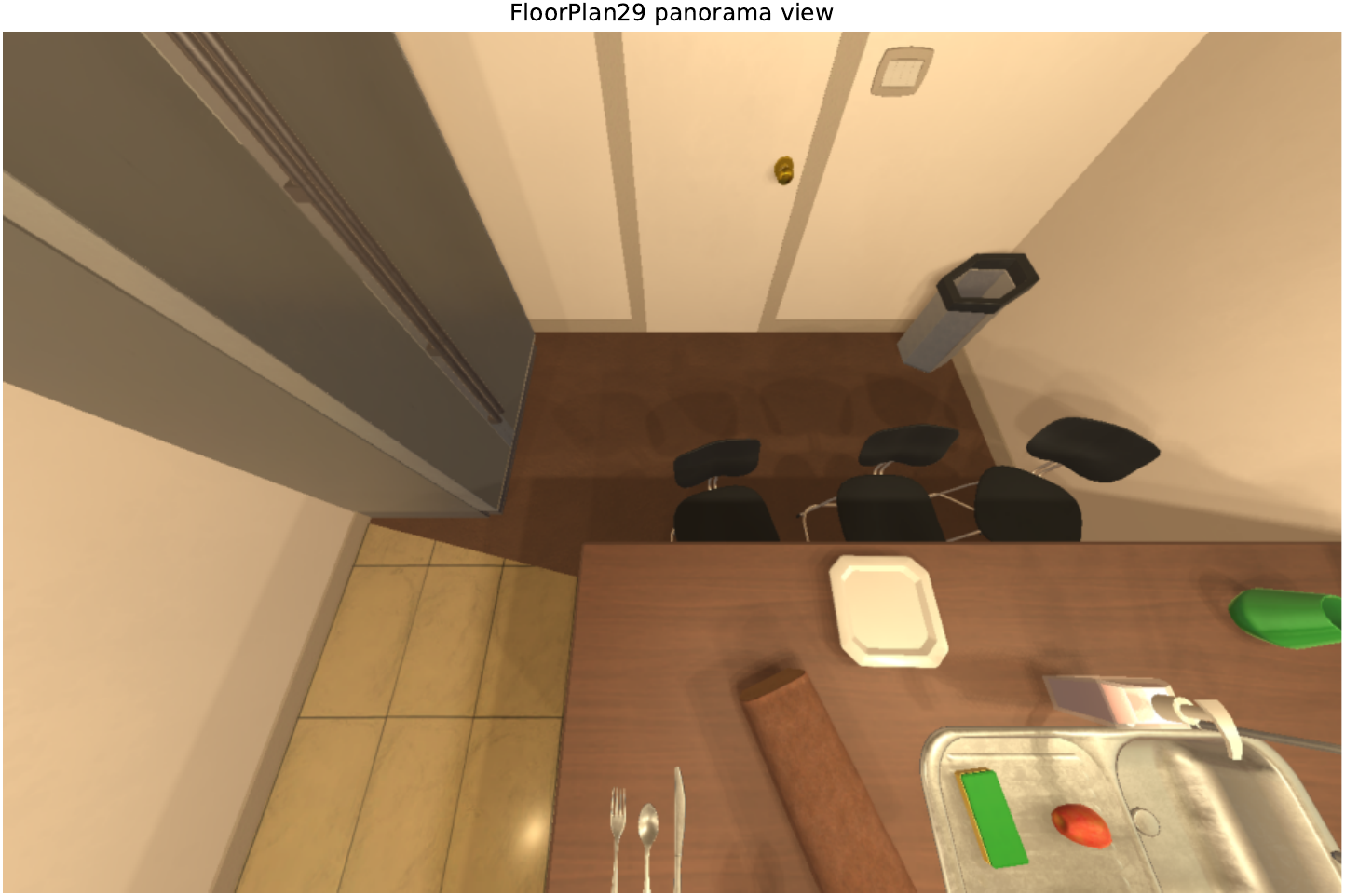}\\[-0.2em]
  {\small \texttt{FloorPlan29}}
\end{minipage}
\end{tabular}
\caption{Examples of AI2-THOR kitchen scenes used in the household interaction domain. The scenes differ in layout, object placement, visibility, and receptacle configuration.}
\label{fig:ai2thor_scenes}
\end{figure}

\begin{table}[t]
\centering
\small
\caption{AI2-THOR simulator configuration.}
\label{tab:ai2thor_sim_config}
\begin{tabular}{ll}
\toprule
Quantity & Value \\
\midrule
Scenes & \texttt{FloorPlan1}--\texttt{FloorPlan29} \\
Image width & \(900\) \\
Image height & \(600\) \\
Visibility distance & \(1.5\) m \\
Field of view & \(90^\circ\) \\
Grid size & \(0.25\) m \\
Maximum episode length & \(250\) steps \\
Depth image & disabled \\
Instance segmentation & disabled \\
\bottomrule
\end{tabular}
\end{table}

\subsection{AI2-THOR Code-as-Policy Interface}
\label{app:ai2thor_policy_interface}

Code as policy evolution generates a Python class named \texttt{Policy}. The class is instantiated as \texttt{Policy(seed=0)}. Before each
rollout, \texttt{reset()} clears episode-level memory. At each environment step, \texttt{compute\_action(obs)} is called once and returns
\[
(a_t,\iota_t),
\]
where \(a_t\) is an action name and \(\iota_t\) is either an AI2-THOR object identifier or \texttt{None}. Navigation and camera actions use \(\iota_t=\texttt{None}\). Object-interaction actions require an object identifier that is valid for the current observation.

\begin{table}[t]
\centering
\small
\caption{AI2-THOR action space exposed to generated policy code.}
\label{tab:ai2thor_actions}
\begin{tabular}{lll}
\toprule
Policy action & Object id required & AI2-THOR action \\
\midrule
\texttt{move\_ahead} & no & \texttt{MoveAhead} \\
\texttt{rotate\_left} & no & \texttt{RotateLeft} \\
\texttt{rotate\_right} & no & \texttt{RotateRight} \\
\texttt{look\_up} & no & \texttt{LookUp} \\
\texttt{look\_down} & no & \texttt{LookDown} \\
\texttt{open\_object} & yes & \texttt{OpenObject} \\
\texttt{close\_object} & yes & \texttt{CloseObject} \\
\texttt{pickup\_object} & yes & \texttt{PickupObject} \\
\texttt{put\_object} & yes & \texttt{PutObject} \\
\texttt{toggle\_on} & yes & \texttt{ToggleObjectOn} \\
\texttt{toggle\_off} & yes & \texttt{ToggleObjectOff} \\
\texttt{done} & no & end episode \\
\bottomrule
\end{tabular}
\end{table}

For every object-parameterized action, the returned identifier must belong to the corresponding valid-action list:
\[
\iota_t \in \mathrm{obs}[\texttt{"action\_space"}][a_t].
\]
The valid object identifiers are recomputed at every step from the simulator state, including visibility, distance, inventory state, object state, and AI2-THOR action preconditions. 
\subsection{Observation Space}
\label{app:ai2thor_obs}

The observation is a Python dictionary containing the current task, scene, egocentric agent state, inventory state, task-object fields, visible object summaries, task-progress predicates, valid actions, the previous action result, and step counters.

\begin{table}[t]
\centering
\small
\caption{Top-level AI2-THOR observation fields.}
\label{tab:ai2thor_obs_top}
\begin{tabular}{ll}
\toprule
Field & Content \\
\midrule
\texttt{task} & Task name \\
\texttt{task\_description} & Natural-language task description \\
\texttt{scene} & Current AI2-THOR scene id \\
\texttt{scene\_is\_fixed} & Boolean scene flag \\
\texttt{agent} & Agent position, rotation, and camera horizon \\
\texttt{inventory} & Held object type and held object identifier \\
\texttt{targets} & Existence, identifier, and visibility fields for task objects \\
\texttt{visible\_objects} & Summaries of currently visible objects \\
\texttt{task\_progress} & Boolean task-progress predicates \\
\texttt{available\_actions} & List of action names \\
\texttt{action\_space} & Valid object identifiers for object actions \\
\texttt{last\_action\_result} & Previous action, success flag, and failure information \\
\texttt{step\_count} & Current episode step index \\
\texttt{max\_steps} & Episode step limit \\
\bottomrule
\end{tabular}
\end{table}

The \texttt{agent} field contains position, rotation, and camera horizon. The \texttt{inventory} field contains the held object type and held
object identifier. The \texttt{targets} field reports object existence, object identifier, and visibility for Apple, Bread, Microwave, and
Fridge:
\[
\begin{aligned}
\texttt{targets}=\{&
\texttt{apple\_exists\_in\_scene},
\texttt{apple\_object\_id},
\texttt{apple\_visible},\\
&
\texttt{bread\_exists\_in\_scene},
\texttt{bread\_object\_id},
\texttt{bread\_visible},\\
&
\texttt{microwave\_exists\_in\_scene},
\texttt{microwave\_object\_id},
\texttt{microwave\_visible},\\
&
\texttt{fridge\_exists\_in\_scene},
\texttt{fridge\_object\_id},
\texttt{fridge\_visible}
\}.
\end{aligned}
\]

\begin{table}[t]
\centering
\small
\caption{Task-progress predicates for the AI2-THOR household task.}
\label{tab:ai2thor_progress}
\begin{tabular}{ll}
\toprule
Predicate & Meaning \\
\midrule
\texttt{apple\_picked} & Apple is currently held \\
\texttt{bread\_picked} & Bread is currently held \\
\texttt{microwave\_open} & Microwave is open \\
\texttt{fridge\_open} & Fridge is open \\
\texttt{apple\_in\_microwave} & Apple is inside the Microwave \\
\texttt{bread\_in\_fridge} & Bread is inside the Fridge \\
\texttt{microwave\_on} & Microwave is toggled on \\
\texttt{success} & Full task success predicate is satisfied \\
\bottomrule
\end{tabular}
\end{table}

Each entry in \texttt{visible\_objects} is a summary of one currently visible object. These summaries expose the object type, exact object identifier, visibility, distance, interaction affordances, current object state, and containing receptacles.

\begin{table}[t]
\centering
\small
\caption{Fields in each visible object summary.}
\label{tab:ai2thor_visible_object}
\begin{tabular}{ll}
\toprule
Field & Content \\
\midrule
\texttt{objectType} & Semantic object category \\
\texttt{objectId} & Exact identifier required for object actions \\
\texttt{name} & AI2-THOR object name when available \\
\texttt{visible} & Visibility flag \\
\texttt{distance} & Distance from the agent when available \\
\texttt{pickupable} & Whether the object can be picked up \\
\texttt{openable} & Whether the object can be opened or closed \\
\texttt{isOpen} & Current open state \\
\texttt{receptacle} & Whether the object can receive other objects \\
\texttt{toggleable} & Whether the object can be toggled \\
\texttt{isToggled} & Current toggle state \\
\texttt{isPickedUp} & Whether the object is currently picked up \\
\texttt{parentReceptacles} & Current containing receptacles, when available \\
\bottomrule
\end{tabular}
\end{table}

The previous action result is part of the observation. It contains the previous action name, object identifier, success flag, normalized error category, and raw error message:
\[
\texttt{last\_action\_result}
=
\{
\texttt{action},
\texttt{object\_id},
\texttt{success},
\texttt{error\_category},
\texttt{error\_message}
\}.
\]
The error category can indicate blocked movement, collision, a non-visible object, a non-reachable object, a failed precondition, a generic failure, or a simulator exception.

\subsection{Partial Observability and Object Identifiers}
\label{app:ai2thor_partial_obs}

The policy does not receive a complete scene graph at every step. \texttt{visible\_objects} contains only objects visible from the current egocentric pose. Task-object identifiers may therefore be unavailable until the agent changes viewpoint, moves closer, or opens a receptacle. The exact \texttt{objectId} string is required for object actions, and object identifiers can differ across floor plans.

The action space is the source of validity for interactions. Even when a target object identifier is known from memory, an interaction is valid only if that identifier appears in the corresponding list in \texttt{obs["action\_space"]}. This makes the policy interface state-dependent: the set of valid object actions changes with visibility, distance, object state, inventory state, and simulator preconditions.

\section{Additional Results}
\label{app:additional_results}

\subsection{Evaluator Evolution Results}
\label{app:evaluator_evolution_results}

Section~\ref{subsec:evaluator_evolution} defines evaluator evolution and the fixed evaluator \(E^\star\) used during code-as-policy search. Here, we report the task-specific form of the selected evaluators. In both domains, \(E^\star\) returns scalar fitness for selection and feedback
metrics for subsequent code-as-policy proposals:
\[
E^\star(\mathcal{T}_{N_{\mathrm{roll}}}(p)) = (F(p), \mathbf{M}(p)).
\]
The empirical success rate \(S(p)\) is computed from the task success indicator and recorded alongside.

\subsubsection{Robosuite Franka Tower-of-Hanoi}
\label{app:franka_evaluator_evolution_results}

For Robosuite, the selected evaluator is defined for the four-cube Tower-of-Hanoi task with cubes
\[
\mathcal{K}=
\{\mathrm{xlarge},\mathrm{large},\mathrm{medium},\mathrm{small}\}
\]
and containers \(A,B,C\). Let \(\mathrm{in}_j(c)\) denote that cube \(c\) lies inside the hole region of container \(j\), and let \(\mathrm{stack}(c_1,c_2)\) denote that cube \(c_2\) is stacked on cube \(c_1\) with the required horizontal and vertical tolerances. The binary success predicate requires all cubes to be in the target container \(C\), containers \(A\) and \(B\) to be empty, and the stack on \(C\) to satisfy the legal Hanoi order:
\[
\begin{aligned}
\mathrm{success}_e ={}&
\left(\bigwedge_{c\in\mathcal{K}} \mathrm{in}_C(c)\right)
\wedge
\left(\bigwedge_{c\in\mathcal{K}}
\neg \mathrm{in}_A(c)\wedge \neg \mathrm{in}_B(c)\right) \\
&\wedge\;
\mathrm{stack}(\mathrm{xlarge},\mathrm{large})
\wedge
\mathrm{stack}(\mathrm{large},\mathrm{medium})
\wedge
\mathrm{stack}(\mathrm{medium},\mathrm{small}) .
\end{aligned}
\]
The task success indicator is
\[
S_e=\mathbb{I}\{\mathrm{success}_e\},
\]
computed from the environment success predicate. The evaluator also returns a dense episode fitness \(f_e\in[0,1]\). The policy-level fitness is the mean episode fitness over the rollout budget:
\[
F(p)=\frac{1}{N_{\mathrm{roll}}}
\sum_{e=1}^{N_{\mathrm{roll}}} f_e(p).
\]

At each control step \(t\), the evaluator computes phase scores for approach, grasp/lift, transport, placement, and final arrangement. The approach score is the maximum proximity score between the end effector and any cube:
\[
A_t=\max_{c\in\mathcal{K}}
\exp\!\left(-\frac{\|x_{\mathrm{eef},t}-x_{c,t}\|}{0.12}\right).
\]
The grasp/lift score is the maximum normalized lift height:
\[
G_t=\max_{c\in\mathcal{K}}
\mathrm{clip}_{[0,1]}
\left(
\frac{z_{c,t}-(z_{\mathrm{table}}+s_c)}{0.08}
\right),
\]
where \(s_c\) is the cube half-edge length. The transport score rewards lifting a cube while moving it toward the target container:
\[
T_t=\max_{c\in\mathcal{K}}
\ell_{c,t}
\exp\!\left(-\frac{d_C(c,t)}{0.18}\right),
\]
where \(\ell_{c,t}\) is the normalized lift score of cube \(c\), and \(d_C(c,t)\) is the planar error between cube \(c\) and the target hole
center in container \(C\).

Placement is measured using continuous hole scores. Let \(h_j(c,t)\in[0,1]\) be the score that cube \(c\) lies inside the hole of
container \(j\). The target-fill and source-emptiness terms are
\[
C_{\mathrm{fill},t}
=
\frac{1}{4}\sum_{c\in\mathcal{K}} h_C(c,t),
\]
\[
E_{AB,t}
=
\frac{1}{2}
\left(
1-\frac{1}{4}\sum_{c\in\mathcal{K}}h_A(c,t)
+
1-\frac{1}{4}\sum_{c\in\mathcal{K}}h_B(c,t)
\right).
\]
The placement score is
\[
L_t=
\mathrm{clip}_{[0,1]}
\left(
0.75\,C_{\mathrm{fill},t}
+
0.25\,E_{AB,t}
\right).
\]

The arrangement score evaluates the target stack on container \(C\):
\[
R_t =
\mathrm{clip}_{[0,1]}
\left(
0.50\,\bar{g}_t
+
\frac{1}{6}\sigma_{\mathrm{xlarge},\mathrm{large},t}
+
\frac{1}{6}\sigma_{\mathrm{large},\mathrm{medium},t}
+
\frac{1}{6}\sigma_{\mathrm{medium},\mathrm{small},t}
\right),
\]
where \(\bar{g}_t\) is the mean goal-pose score over the four cubes and \(\sigma_{i,j,t}\) is the continuous score that cube \(j\) is stacked on
cube \(i\).

The per-step composite score is
\[
\tilde{P}_t =
0.14\,A_t
+0.18\,G_t
+0.20\,T_t
+0.20\,L_t
+0.28\,R_t .
\]
Illegal Hanoi placements are penalized by the maximum continuous score for a larger cube being placed on a smaller cube:
\[
P_t =
\mathrm{clip}_{[0,1]}
\left(
0.95\,\tilde{P}_t(1-0.15\,U_t)
+
0.05\,Q_{\mathrm{int},t}
\right),
\]
where \(U_t\in[0,1]\) is the illegal-stack score and \(Q_{\mathrm{int},t}\) measures interaction smoothness from end-effector and cube speeds.

For an episode ending at step \(T_e\), the evaluator records the time-average composite score, the best composite score, and the final arrangement score:
\[
P_{\mathrm{auc}}
=
\frac{1}{T_e}\sum_{t=1}^{T_e}P_t,
\qquad
P_{\mathrm{best}}
=
\max_{1\leq t\leq T_e}P_t,
\qquad
R_{\mathrm{final}}
=
R_{T_e}.
\]
The episode fitness is
\[
f_e =
\mathrm{clip}_{[0,1]}
\left(
0.45\,P_{\mathrm{auc}}
+
0.45\,P_{\mathrm{best}}
+
0.10\,R_{\mathrm{final}}
\right).
\]
If the Tower-of-Hanoi success predicate is satisfied, \(f_e\) is set to \(1.0\).

The feedback metrics returned by the evaluator include maximum and final phase scores for approach, grasp/lift, transport, placement, and arrangement; milestone times for first reach, first lift, first cube in \(C\), first two cubes in \(C\), and all cubes in \(C\); minimum end-effector--cube distances; maximum lift heights; cube transport speeds; hole scores for containers \(A\), \(B\), and \(C\); target-fill and source-emptiness scores; goal-pose scores; legal stack-pair scores; drop count; and maximum illegal-stack score. These feedback metrics localize failure to reaching, lifting, transport, placement, final stack arrangement, illegal Hanoi placement, or unstable interaction, and are
included with scalar fitness when generating subsequent code-as-policy proposals.

\subsubsection{AI2-THOR Household Interaction}
\label{app:ai2thor_evaluator_evolution_results}

For AI2-THOR, the selected evaluator is defined for the task \texttt{apple\_in\_microwave\_on\_bread\_in\_fridge}. The binary success predicate is the same as in Appendix~\ref{app:ai2thor_task}:
\[
\mathrm{success}_e =
\mathrm{apple\_in\_microwave}
\;\wedge\;
\mathrm{microwave\_on}
\;\wedge\;
\mathrm{bread\_in\_fridge}.
\]
The task success indicator is
\[
S_e=\mathbb{I}\{\mathrm{success}_e\},
\]
computed from the environment success predicate.
The evaluator returns a dense episode fitness \(f_e\in[0,1]\). The policy-level fitness is the mean episode fitness over the rollout budget:
\[
F(p)=\frac{1}{N_{\mathrm{roll}}}\sum_{e=1}^{N_{\mathrm{roll}}} f_e(p).
\]

At each environment step \(t\), the evaluator computes a phase-progress score \(P_t\) from the current task predicates and accumulated visibility history:
\[
\begin{aligned}
P_t ={}&
0.06\,p_{\mathrm{apple\_seen},t}
+0.06\,p_{\mathrm{bread\_seen},t}
+0.08\,p_{\mathrm{apple\_picked},t}
+0.08\,p_{\mathrm{bread\_picked},t} \\
&+0.10\,p_{\mathrm{microwave\_seen},t}
+0.10\,p_{\mathrm{fridge\_seen},t}
+0.08\,p_{\mathrm{microwave\_open},t}
+0.08\,p_{\mathrm{fridge\_open},t} \\
&+0.16\,p_{\mathrm{apple\_in\_microwave},t}
+0.16\,p_{\mathrm{bread\_in\_fridge},t}
+0.04\,p_{\mathrm{microwave\_on},t}.
\end{aligned}
\]
Visibility terms are fractional and depend on whether the object has been observed and, when visible, its distance to the agent. State-change terms such as pickup, receptacle opening, object insertion, and toggling are binary once the corresponding predicate is reached.

For an episode ending at step \(T_e\), the evaluator records
\[
P_{\mathrm{best}}=\max_{0\leq t\leq T_e}P_t,
\qquad
P_{\mathrm{final}}=P_{T_e}.
\]
The progress term used in the episode fitness is
\[
P = 0.35\,P_{\mathrm{best}} + 0.65\,P_{\mathrm{final}} .
\]
Thus, \(P_{\mathrm{best}}\) gives credit for the best intermediate task state reached during the rollout, while \(P_{\mathrm{final}}\) measures the task progress that remains at termination.

The final episode fitness is
\[
\begin{aligned}
f_e = \mathrm{clip}_{[0,1]}\big(
&0.74\,P
+0.14\,Q_{\mathrm{int}}
+0.06\,Q_{\mathrm{vis}}
+0.06\,Q_{\mathrm{eff}} \\
&-0.07\,C_{\mathrm{blocked}}
-0.03\,C_{\mathrm{carry}}
\big).
\end{aligned}
\]
Here \(Q_{\mathrm{int}}\) measures valid and task-relevant object interactions, \(Q_{\mathrm{vis}}\) measures visibility of the Apple, Bread, Microwave, and Fridge, and \(Q_{\mathrm{eff}}\) rewards earlier progress within the episode horizon. The penalty \(C_{\mathrm{blocked}}\) increases with failed movement and long intervals without progress. The penalty \(C_{\mathrm{carry}}\) increases when the policy holds the Apple or Bread for many steps without placing it in the required receptacle.

The evaluator also enforces lower bounds on the milestones of \(f_e\). Any successful Apple or Bread pickup provides at least \(0.06\) fitness. Observing the Microwave or Fridge gives at least \(0.10\). Opening the Microwave or Fridge gives at least \(0.18\). Placing either the Apple in the Microwave or the Bread in the Fridge gives at least \(0.42\). Completing both placement predicates gives at least \(0.72\). Completing both placement predicates and turning on the Microwave gives at least \(0.90\). Full task success gives at least \(0.98\).

The feedback metrics returned by the evaluator include terminal task predicates, best- and final-phase scores, first-hit times for task milestones, visibility and proximity statistics for all target objects, action success rates, task-relevant interaction success rates, failed movement rate, and carrying-stall measures. The milestone times record when the policy first picks up the Apple or Bread, first observes the Microwave or Fridge, opens each receptacle, places each object in its target receptacle, and toggles the Microwave on. These feedback metrics identify the unsatisfied AI2-THOR precondition at the level used by the code-as-policy interface: object discovery, valid pickup, receptacle localization, receptacle opening, object placement, Microwave toggling, or premature termination. 

\subsection{Evolved Code-as-Policy Programs}
\label{app:best_evolved_policies}

We report one successful evolved code-as-policy program from each domain to illustrate the executable strategy produced by policy evolution. Since multiple final policies solve the task, the discussion focuses on the policy structure: task-state representation, control-flow logic, closed-loop execution, and recovery mechanisms.

\subsubsection{Robosuite Franka Tower-of-Hanoi}
\label{app:best_evolved_franka_policy}

The Franka code-as-policy program combines a fixed symbolic Hanoi schedule with guarded closed-loop execution. The symbolic component generates the recursive \(15\)-move solution for four cubes and three containers. The policy state is
\[
(i,z,r),
\]
where \(i\) is the current Hanoi move index, \(z\) is the manipulation state, and \(r\) is the retry index. The move index is advanced only after the corresponding physical placement passes a post-placement check.

At each control step, the policy filters the observed end-effector pose, cube poses, and container poses using separate exponential moving averages. It then computes the three container hole centers from the current container poses. For the active Hanoi move, the pick target is the observed pose of the selected cube plus a retry-dependent planar offset. The target place is the observed destination hole center with a height determined from the current top surface of the destination stack, excluding the cube being moved.

The manipulation controller is implemented as a finite-state machine with pick, grasp-verification, placement, centering, release, and retreat states. In the pick stage, the policy moves to a travel height above the selected cube, descends to a pre-grasp pose, and closes the gripper. In the grasp-verification stage, the policy lifts the cube and checks whether it has moved upward from its grasp-time pose while remaining close to the end effector. In the placement stage, the policy moves above the destination container and descends only once it is horizontally aligned. In the centering and release states, the policy centers on the target hole at contact height, then opens the gripper. After release, the policy
checks whether the cube is near the destination hole before advancing the move index.

Recovery is local to the active Hanoi move. If grasp verification fails, the gripper is opened, the retry index is incremented, and the same Hanoi move is attempted again with a different planar pick offset. If the post-placement check fails, the move index is not advanced, so the same Hanoi move is retried from the updated observation. Thus, the symbolic Hanoi ordering remains fixed, but the execution of each move is adapted through retry-conditioned targets.

The controller also uses contact-aware motion constraints. Downward motion is suppressed until the horizontal error is below the corresponding pick or place gate. Cartesian commands are slew-limited, and lateral motion is reduced near the stack to avoid disturbing previously placed cubes. Before release, a centering state applies small lateral corrections until the residual planar error and commanded lateral motion are both small. The largest cube uses a tighter placement gate and a longer centering window, since its placement forms the base of the final tower.

\subsubsection{AI2-THOR Household Interaction}
\label{app:best_evolved_ai2thor_policy}

The selected AI2-THOR code-as-policy program solves the task by using observation-conditioned object interactions rather than a fixed action sequence. The policy maintains memory of object identifiers and parsed object positions for the Apple, Bread, Microwave, and Fridge. At each step, it selects object identifiers only from the current \texttt{obs["action\_space"]}, so pickup, open, put, and toggle actions are issued only when the simulator exposes them as executable.

Navigation uses the agent grid pose and remembered object positions. The policy records failed \texttt{move\_ahead} attempts as blocked grid edges and avoids them in subsequent navigation. When direct movement toward a target object is obstructed, it switches to an axis-aligned greedy choice over neighboring grid cells, using blocked-edge memory and container-specific bad-placement cells to avoid repeating failed poses.

The main domain-specific recovery mechanism is for receptacle placement. If \texttt{put\_object} fails at the Microwave or Fridge, the policy marks the current grid cell as a bad placement pose for that receptacle and executes a short pose-adjustment sequence before retrying the placement from a different viewpoint. Separate recovery queues are used for Microwave and Fridge placement failures, reflecting the different geometry and visibility constraints of the two receptacles.

The policy also handles the hidden-Apple case. If the Apple is not found during a normal search, the policy probes the Fridge, opens it when a valid Fridge identifier is available, scans the open receptacle with camera actions, and attempts to pick up the Apple from inside. If the Apple is picked from an open Fridge, the policy closes the Fridge before navigating to the Microwave, avoiding collisions and failed navigation caused by the open door.

\end{document}